\documentclass[10pt,journal,compsoc]{IEEEtran}
\ifCLASSOPTIONcompsoc
  \usepackage[nocompress]{cite}
\else
  \usepackage{cite}
\fi
\ifCLASSINFOpdf
\else
\fi

\usepackage{amsmath,amsfonts}
\usepackage{algorithmicx}
\usepackage{algorithm}  
\usepackage{algpseudocode}  
\usepackage{amsmath}  
\usepackage{array}
\usepackage{textcomp}
\usepackage{booktabs}
\usepackage{stfloats}
\usepackage{url}
\usepackage{verbatim}
\usepackage{bbding}
\usepackage{graphicx} 
\usepackage{xcolor}

\begin{document}
%
\title{PhysMLE: Generalizable and Priors-Inclusive Multi-task Remote Physiological Measurement}
%
%
%
%

\author{Jiyao Wang,~\IEEEmembership{Student Member,~IEEE,}
        Hao Lu, Ange Wang, Xiao Yang, Yingcong Chen, Dengbo He, and~Kaishun Wu,~\IEEEmembership{Fellow,~IEEE}
\IEEEcompsocitemizethanks{\IEEEcompsocthanksitem This work was supported by the National Natural Science Foundation of China (No. 52202425), Guangzhou Municipal Science and Technology Project (No. 2023A03J0011), and Guangzhou Science and Technology Program City-University Joint Funding Project (No. 2023A03J0001) (Corresponding author: Dengbo He).
\IEEEcompsocthanksitem Jiyao Wang, Ange Wang, and Dengbo He is with the Systems Hub, the Hong Kong University of Science and Technology (Guangzhou), Guangzhou, China,
(e-mail: jwanggo@connect.ust.hk; awang324@connect.hkustgz.edu.cn; dengbohe@ust.hk). Hao Lu, Yingcong Chen, and Kaishun Wu are with the Information Hub, the Hong Kong University of Science and Technology (Guangzhou), Guangzhou, China,
(e-mail: hlu585@connect.ust.hk; yingcongchen@ust.hk; wuks@hkust-gz.edu.cn). Xiao Yang is with Sichuan Agricultural University, Yaan, China,
(e-mail: 202005537@stu.sicau.edu.cn).}

\thanks{Manuscript received May 10, 2024.}}

%
%

\markboth{Journal of \LaTeX\ Class Files,~Vol.~14, No.~8, August~2015}%
{Shell \MakeLowercase{\textit{et al.}}: Bare Demo of IEEEtran.cls for Computer Society Journals}
%



\IEEEtitleabstractindextext{%
\begin{abstract}
Remote photoplethysmography (rPPG) has been widely applied to measure heart rate from face videos. To increase the generalizability of the algorithms, domain generalization (DG) attracted increasing attention in rPPG. However, when rPPG is extended to simultaneously measure more vital signs (e.g., respiration and blood oxygen saturation), achieving generalizability brings new challenges. Although partial features shared among different physiological signals can benefit multi-task learning, the sparse and imbalanced target label space brings the seesaw effect over task-specific feature learning. To resolve this problem, we designed an end-to-end Mixture of Low-rank Experts for multi-task remote Physiological measurement (PhysMLE), which is based on multiple low-rank experts with a novel router mechanism, thereby enabling the model to adeptly handle both specifications and correlations within tasks. Additionally, we introduced prior knowledge from physiology among tasks to overcome the imbalance of label space under real-world multi-task physiological measurement. For fair and comprehensive evaluations, this paper proposed a large-scale multi-task generalization benchmark, named Multi-Source Synsemantic Domain Generalization (MSSDG) protocol. Extensive experiments with MSSDG and intra-dataset have shown the effectiveness and efficiency of PhysMLE. In addition, a new dataset was collected and made publicly available to meet the needs of the MSSDG.
\end{abstract}

}

\maketitle

\IEEEdisplaynontitleabstractindextext

%
\IEEEpeerreviewmaketitle

\IEEEraisesectionheading{\section{Introduction}\label{sec:introduction}}
\IEEEPARstart{P}{hysiological} monitoring plays a vital role in assessing well-being and performance in various sectors, including healthcare, emotional analysis, and driver monitoring systems \cite{gouveia2023low,wu2023recognizing}. Key indicators including heart rate (HR), heart rate variability (HRV), respiration rate (RR), and blood oxygen saturation (SpO2) are crucial for assessing well-being and performance of human subjects. Traditional methods like electrocardiography (ECG) and photoplethysmography (PPG) require direct contact of sensors and skin, causing inconvenience and discomfort for long-term monitoring \cite{akamatsu2023calibrationphys}. To address this problem, non-contact video-based physiological measurement techniques have gained increasing interest. Remote photoplethysmography (rPPG) is highlighted as a promising approach, which can be realized through ubiquitous imaging devices like RGB webcams and smartphone cameras. rPPG methods analyze subtle facial skin periodic light absorption caused by cardiac activities \cite{yu2021facial} to obtain blood volume pulse (BVP) signals, which can further be used to extract HR and HRV \cite{niu2020video}. Previous studies have also attempted to estimate RR \cite{fiedler2020fusion,pourbemany2021real,siam2020efficient} and SpO2 \cite{wu2023peripheral,tarassenko2014non,kim2021non} from RGB facial videos by detecting color changes and movements associated with the blood flow during the breathing cycle and the varying absorption of light by oxygenated and deoxygenated hemoglobin in the blood vessels of the face, respectively. Recently, advancements in deep learning (DL) have shown superior performance in complex scenarios \cite{wu2023peripheral,pourbemany2021real,shao2023tranphys}, making DL-based methods highly promising for non-contact physiological measurement.

Conventional methods of remote physiological measurement are typically concentrated on one predictive task (e.g., BVP or RR), within a single domain \cite{das2021bvpnet,du2021weakly}. These methods are developed by training on datasets gathered from one specific scenario and are tailored to predict one physiological indicator, while is difficult to meet the needs of applications in real scenarios \cite{orphanidou2019review}. For example, patient monitoring systems that rely on a single vital sign may lead to 'alarm fatigue' and further make clinical staff desensitized \cite{Orphanidou2015SignalQualityIF}. At the same time, training separate models for each task incurs high deployment costs and reduces iterative efficiency \cite{chang2023pepnet}. In remote physiological measurement, different targets are functionally related and there are dependencies between multiple tasks. For instance, the hemodynamic changes brought by respiration can influence the blood flow and blood volume in the peripheral vasculature, leading to variations in the BVP signal \cite{peper2007there}. Some studies tried to develop one united model to simutanously estimate multiple vital signs \cite{liu2020multi,narayanswamy2024bigsmall}. However, although these methods utilized the dependencies between different tasks, they neglected the multi-source domain generalization (MSDG) \cite{lu2023neuron}.  

Limited by the laboratory dataset acquisition process, DL methods trained on multi-source datasets suffer from severe performance degradation in the unknown target domain due to variations between different multiple datasets, which was usually defined as the domain shift \cite{wang2023hierarchical}. The domain shift can be caused by different data acquisition equipment, individual differences, and environmental changes. In recent years, to enhance the generalizability of DL-based remote physiological measurement models in the presence of domain shift, MSDG methods for rPPG continuously gained attention in the rPPG field \cite{wang2023hierarchical,10.1145/3581783.3612265,lu2023neuron}. However, previous multi-task methods failed to leverage the full potential of the available data and overlook the semantic differences in the feature space under multi-domain settings, which can result in suboptimal performance \cite{chang2023pepnet}. Further, past MSDG methods focused on aligning the feature semantics under different domains but ignored target dependencies in the label space under multi-task settings. Particularly, in the field of remote physiological measurement, different public datasets only provided partial vital signs labels (see Table \ref{t0}). Thus, we cannot simply and directly reuse MSDG for rPPG to multi-task settings.

\begin{figure}
\begin{center}
\includegraphics[scale=0.29]{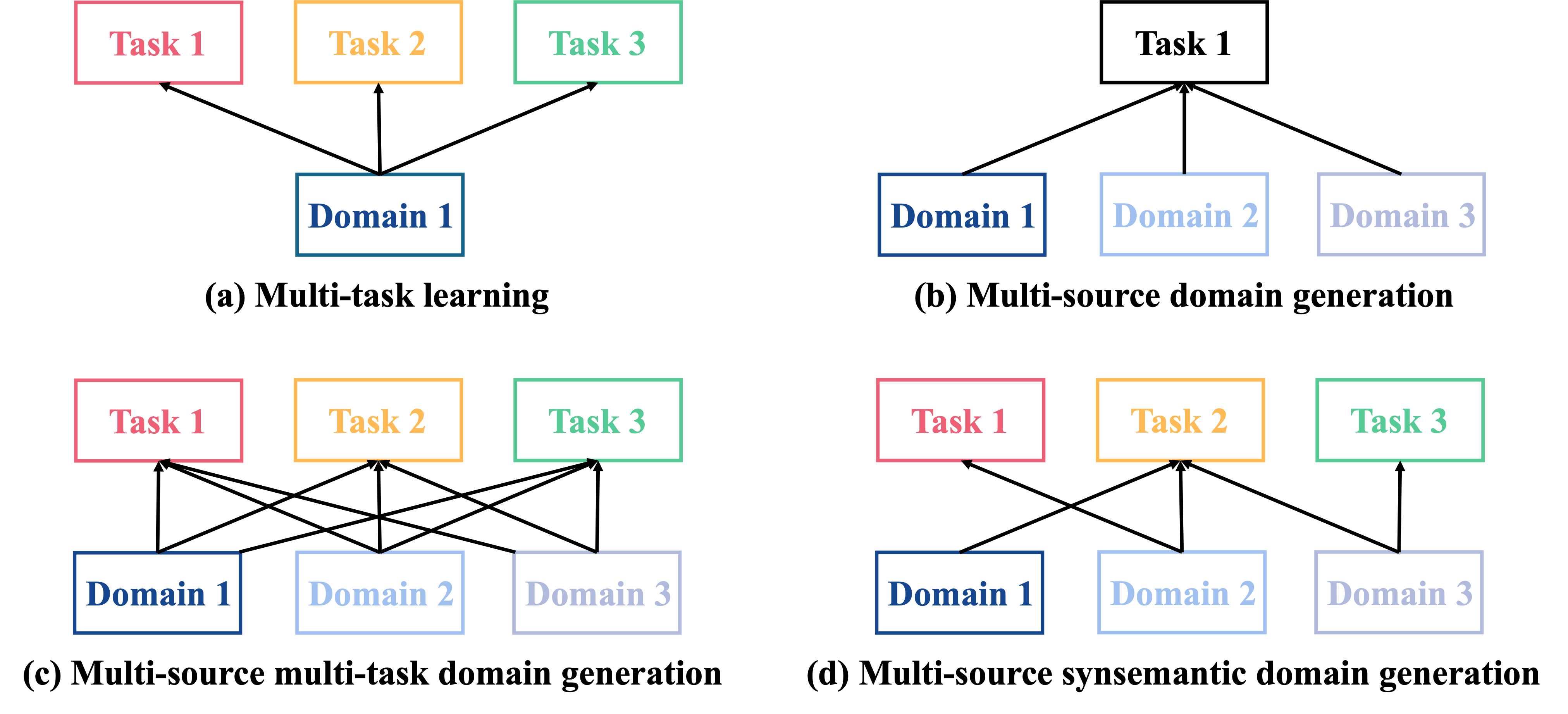}\\
\end{center}
\vspace{-6mm}
\caption{Illustration of the difference between Multi-source Synsemantic Domain Generalization (MSSDG) and classic multi-task learning or multi-source domain generalization.}\label{f1}
\vspace{-3mm}
\end{figure}

Given this, we raise a new challenge of Multi-Source Synsemantic Domain Generalization (MSSDG), which extends beyond the conventional MSDG by incorporating the complexity encountered in real-world applications (as shown in Figure \ref{f1}). Unlike isolated multi-task learning paradigms or traditional MSDG approaches, MSSDG aims to address the simultaneous occurrence of multiple tasks and domains, which represents a more intricate challenge. Specifically, MSSDG must reconcile disparities in feature semantics and their relative significance that arise both across different domains of a singular task and within a single domain of disparate tasks. Additionally, it must account for the varying degrees of target sparsity and the interdependencies that are present among different tasks within the same domain as well as for the same task across various domains.

To address this challenge, inspired by the standout performance in multi-task learning of Mixture-of-Experts (MoE) \cite{shazeer2016outrageously}, we propose an end-to-end Mixture of Low-rank Experts for multi-task remote Physiological measurement (\textbf{PhysMLE}), which fully exploits the dependencies between tasks via constructing multiple low-rank trainable experts. Different from classic MoE architecture, the introduced low-rank expert can significantly reduce the high tuning costs of MoE, and implement plug-and-play. Besides, the proposed PhysMLE can also address the limitations of previous Parameter Efficient Fine-Tuning (PEFT) methods (e.g., P-tuning \cite{liu2023gpt}, LoRA \cite{hu2021lora}) when faced with multi-task setting. Specifically, PEFT methods need to be fine-tuned for multiple sets of parameters for each task separately, or one set of parameters for all tasks. In brief, PhysMLE takes the primary decoded feature from the original first block of the substitutable backbone network as input. Then, we replaced the core feature learning module of the backbone with our PhysMLE layer, which consists of multiple learnable Low-rank Adaptation (LoRA) experts, multiple expert-specific feature routers, and the original frozen weights. During training, a regularization based on weight orthogonality is proposed to force each expert to learn distinct information to avoid overlapping of different physiological features. The expert-specific feature routers are designed to perform element-wise selection on the generated feature from each LoRA expert. After processing through multiple-layer PhysMLE basic blocks, the features are fed to each task-specific router separately for gating to obtain a sparse task-specific representation, which is further used for task-specific head for sign estimation.

Moreover, to cope with domain shift in MSSDG with unbalanced target task space, we propose the following solution. In the MSSDG setting, the challenge of generalizability exists not only in the conditional domain shift \cite{Zhang2013DomainAU} due to environmental changes, but also in identifying the invariance between the inputs and the labeling of the varying tasks in different domains. In the field of remote physiological monitoring based on face videos, inspired by \cite{sun2024}, we argue that the physiological signals reflected from different parts of the face should remain consistent. Also, the changes in a person's physiological characteristics over a short time (5 seconds) should be temporally smooth and gradual. Therefore, we constrain the aggregated representation of each layer of the PhysMLE basic block to remain semantically spatio-temporally consistent through two new regularizations. Furthermore, to resolve the imbalance of target tasks in MSSDG, considering the heterogeneous characteristics and correlations among different physiological signals, we strengthen the learning capacity of PhysMLE through three regularizations based on prior knowledge from biomedical fields \cite{berntson1997heart,tarassenko2014non}. We summarize the contributions of this work as follows:

\begin{itemize}
    \item We proposed a novel end-to-end framework PhysMLE, which adapts the strengths of MoE and LoRA to efficiently learn multiple low-rank expert weights to avoid gradient contradiction in multi-task settings. A weight orthogonality constraint was used to ensure that the widest possible gradient space is learned, and an element-wise router was tailored to a flexible combination of expert features for maximizing the use of positive associations between tasks. 
    
    \item To resolve domain shift and unbalanced target task space under MSSDG, we not only enhance the semantic spatio-temporal consistency in each layer of PhysMLE, but also propose three task-specific priors to augment the imbalance label space, which is introduced in rPPG modeling for the first time.
    \item As the first study to tackle domain shift in multi-task remote physiological measurement, we proposed a new Multi-Source Synsemantic Domain Generalization (MSSDG) protocol. Considering the shortcomings of the previously public dataset, we collected and publicized a new dataset (i.e., High Cognitive Load Workers (HCW)). Jointly with five previously public datasets, we built a new large MSSDG benchmark. Extensive experiments demonstrated the superiority of our method.
\end{itemize}

\begin{table*}[t]
\setlength{\tabcolsep}{1mm}
\centering
\scriptsize
\caption{Summary of Public Non-synthetic Datasets.}
\begin{tabular}{ccccccccccc}
\toprule 
 \textbf{Dataset}    & \textbf{Date} & \textbf{Frames}    & \textbf{Subject} & \textbf{Camera Setting}                    & \textbf{HR} & \textbf{BVP} & \textbf{SpO2} & \textbf{RR} & \textbf{Special Human State} & \textbf{Collection Environment}  \\
\midrule 
MAHNOB-HCI \cite{soleymani2011multimodal} & 2012 & \_        & 27      & Allied Vision Stingray F-046C     & \Checkmark  & \XSolid & \XSolid & \Checkmark                 & Emotion  & $\mid$           \\
\textbf{PURE} \cite{stricker2014non}      & 2014 & 168,120   & 10      & eco274CVGE                        & \Checkmark  & \Checkmark                 & \Checkmark                 & \XSolid & \_            &   $\mid$    \\
COHFACE \cite{heusch2017reproducible}   & 2016 & \_        & 40      & Logitech HD C525                  & \Checkmark  & \XSolid & \XSolid & \Checkmark                 & \_                   & $\mid$\\
MMSE-HR \cite{tulyakov2016self}   & 2016 & \_        & 40      & Di3D                  & \Checkmark  & \XSolid & \XSolid & \XSolid                & Emotion                 & $\mid$ \\
OBF \cite{li2018obf}  & 2018 & \_        & 100     & Blackmagic URFA mini              & \Checkmark  & \XSolid & \XSolid & \Checkmark                 & Atrial fibrillation & Laboratory environment \\
\textbf{UBFC-rPPG} \cite{bobbia2019unsupervised} & 2019 & 57,420    & 42      & Logitech C920 HD Pro              & \Checkmark  & \Checkmark                 & \XSolid & \XSolid & \_                 & $\mid$ \\
\textbf{BUAA}  \cite{xi2020image} & 2020 & 129,549        & 15      & Logitech  C930E HD pro            & \Checkmark  & \Checkmark                 & \XSolid & \XSolid & \_                  & $\mid$\\
\textbf{VIPL-HR} \cite{niu2019rhythmnet} & 2020 & 1,834,785        & 107     & Three cameras & \Checkmark  & \Checkmark                 & \Checkmark                 & \XSolid & \_                 & $\mid$ \\
\textbf{V4V}   \cite{revanur2021first}  & 2021 & 799,113        & 140     & Di3D                              & \Checkmark  & \Checkmark                 & \XSolid & \Checkmark                 & Emotion          & $\mid$  \\
MMPD \cite{tang2023mmpd}   & 2023 & 1,188,000 & 33      & Galaxy S22 Ultra                  & \Checkmark  & \XSolid & \XSolid & \XSolid & \_                 & $\mid$ \\
\midrule 
\textbf{HCW}        & 2024 & 1,565,230 & 41      & 32 laptop webcams    & \Checkmark  & \Checkmark                 & \XSolid & \Checkmark                 & Cognitive workload & Real-world workspace\\
\bottomrule 
\end{tabular}
\textnormal{\\Notes: The datasets we used in this work are bolded in \textbf{black}.}
\label{t0}
\end{table*}

\section{Related Work}
\subsection{Camera-based Physiological Measurement}
Since 2008, among the non-contact techniques, remote photoplethysmography (rPPG) has emerged as a particularly promising approach for physiological measurement \cite{verkruysse2008remote}. Verkruysse et al. \cite{verkruysse2008remote} pointed out that the green channel contains the strongest plethysmography signal because hemoglobin absorbs green light the most, and the red and blue channels also contain plethysmography information. The basic theory is built on optical frameworks such as the Lambert-Beer law (LBL) and Shafer’s dichromatic reflection model (DRM) to model how light interacts with the skin \cite{tarassenko2014non}. Based on this, traditional methods rely on statistical information about heartbeats and often require manual adjustments to filter out inaccurate data \cite{verkruysse2008remote, de2013robust, wang2016algorithmic}. However, these methods often required manual adjustments and struggled with variations in skin tones and lighting conditions. Recently, DL technologies have been developed to improve accuracy in complex environments. Chen and McDuff \cite{chen2018deepphys} first developed a deep model called DeepPhys that extracts BVP signals with convolutional neural networks (CNN) by measuring the difference between frames. In recent years, other advanced methods based on different feature extractors (e.g., 3DCNN \cite{yu2019remote}, Transformer \cite{yu2023physformer++}) were proposed as well. 

Besides HR, RR is also a critical vital sign that is critical for health monitoring. Some early studies remotely measured RR by detecting respiratory motion through pixel movements \cite{cheng2023motion} or pixel intensity variations \cite{janssen2015video, schrumpf2019exploiting}. Some rPPG-based studies adopted traditional methods \cite{ghodratigohar2019remote} or DL \cite{liu2020multi, narayanswamy2024bigsmall} to measure RR. At the same time, existing rPPG works also estimated the frequency domain attributes of predicted BVP signals for RR estimation \cite{li2018obf, niu2020CVD,yu2023physformer++}.
However, existing methods are still troubled by motion artifacts and when the respiration is non-periodic \cite{du2021weakly}, there are errors in estimated BVP, making it challenging for indirect RR inferencing.

Another vital sign, SpO2, has traditionally been measured by analyzing direct current (DC) and alternating current (AC) components at wavelengths of 520 nm and 660 nm, with specialized cameras sensitive to specific wavelengths of light that differentiate oxygenated and deoxygenated hemoglobin \cite{kong2013non}. However, recent efforts aim to estimate SpO2 using readily available RGB cameras by analyzing subtle changes in facial skin color \cite{bal2015non,tarassenko2014non}. Emerging DL-based approaches demonstrate promising results in contactless SpO2 estimation \cite{hu2023contactless,akamatsu2023blood}, although further development is required to achieve clinical-grade accuracy and robustness.

\subsection{Remote Multi-task Physiological Measurement}
The complex multi-task issue was often handled by decomposing them into multiple simple and mutually independent single tasks. However, training models for each task separately incur high deployment costs, and such an approach often ignores the correlation information enriched among tasks. To address these problems, multi-task learning that puts related tasks together has been proposed \cite{caruana1997multitask}. Multi-task learning was widely used in deep learning \cite{ misra2016cross, liu2019end, yang2024multi, shazeer2017outrageously,lu2024gpt}. The MoE model \cite{shazeer2017outrageously} turns the parameters shared by all samples in each stratum into multiple sets of parameters to increase model capacity. However, previous MoE-based methods \cite{ma2018modeling} directly build multiple complete feature encoders as experts, which introduces more computational parameters. Recently, many works attempted to adapt LoRA \cite{hu2021lora} to MoE structure \cite{liu2023moelora, chen2023octavius}. However, they focus mainly on LLM fine-tuning rather than multitasking learning.

In traditional rPPG tasks, the output is usually performed only for a single task (i.e., HR or BVP) \cite{das2021bvpnet,du2021weakly}. Due to different vital signs measured by rPPG are often correlated, recently, MTTS-CAN \cite{liu2020multi} performed cardiovascular and respiratory measurements in real-time, and BigSmall \cite{narayanswamy2024bigsmall} outputs facial movements, respiratory and HR characteristics simultaneously. However, these studies ignored SpO2, which also correlates with HR, RR \cite{yousefi2013motion}. Furthermore, previous multi-task rPPG works were usually trained on a single domain with complete multi-task labels, and since current publicly available datasets often only provide partial physiological signal labels (referring to Table \ref{t0}). Thus, previous MoE or LoRA alone may not resolve the multi-task physiological measurement problem.

\subsection{Domain Generalization in Remote Photoplethysmography}
There are also differences between different domains in rPPG tasks, to address this problem, many DG methods \cite{du2023dual,lu2023neuron,wang2023hierarchical} have been applied to rPPG tasks. The main goal of the DG technique is to generalize the model trained in source domain data to the unseen target domain. The mainstream methods can be divided into three categories: data manipulation, representation learning, and meta-learning. Data manipulation is mainly done through data augmentation methods to increase the number of samples or using data generation methods to produce different samples \cite{shankar2018generalizing,shi2020towards}. Representation learning \cite{ganin2015unsupervised,sun2016return,jiyao2023DGrppg} adapts models to different domains by learning domain-invariant features. Finally, meta-learning approaches \cite{finn2017model,li2018learning,lv2022causality} induces domain shifts in virtual data during training, allowing the model to handle unknown domains better.

For example, NEST \cite{lu2023neuron} improved the generalization ability of the model by maximizing the feature space during training. However, ignoring instance-specific variations tends to affect the results of the model when confronted with specific samples that lack instances during inference. HSRD \cite{wang2023hierarchical} was proposed to measure BVP signals by separating domain-invariant and instance-specific feature space, but it still relied on explicit domain labels for feature learning. Since existing DG methods for remote physiological monitoring are limited to HR and BVP, when DG is applied to the multi-task setting, semantic differences in the feature space and imbalance target tasks might restrict the performance of previous approaches.

\begin{table}[h]
\caption{A summary of symbols and descriptions}\label{t1}
\begin{tabular}{c|l} 
Symbol & Description\\
\hline
$L, W, C$ & Length, width, and the number of channels of a STMaps.\\
$B$ & Batch size.\\
$N$ & Number of tasks.\\
$K$ & Number of low-rank experts.\\
$T$ & Number of PhysMLE layers in the backbone network.\\
$M$ & Number of basic blocks in the backbone network.\\
$E$ & Low-rank expert.\\
$G$ & EFRouter.\\
$\textbf{W}, \textbf{W}_\Delta$ & The frozen and updated weights in the expert.\\
$\textbf{B}, \textbf{A}$ & The low-rank weight matrix in the expert.\\ 
$d_{in}, d_{out}$ & Input and output dimension of weights.\\
$r$ & The number of ranks.\\
$\alpha$ & Hyper-parameter to scale the weights.\\
$\gamma$ & Hyper-parameter to enlarge the selected feature.\\
$s$ & Feature encoded by each PhysMLE basic layer.\\
$s^{'}$ & Task-agnostic feature outputted by the PhysMLE.\\
$s^{''}$ & Task-specific feature after element-wise selection.\\
$X^{'}, Y^{'}$ & Spatial-temporal augmented STMaps and labels.\\
$\hat{Y}$ & Estimated label by PhysMLE.\\
$tau, \mathrm{T}$ & Augmentation ratios for SpO2 prior.\\
$\delta$ & Soft boundary of the difference between estimated RR\\
&  and RR extracted from ground-truth BVP. \\
$p_i$ & Hyper-parameter for joint training. \\
$\mathrm{I}$ & Matrix that all elements are 1.\\
$diag$ & Identity matrix.\\
$\epsilon$ & Random noise sampled from Gaussian distribution.\\
$\mathcal{L}$ & The loss fuction of this work.\\
\end{tabular}\\
\end{table}

\section{Methodology}
This section formally introduces the Mixture of Low-rank Experts for multi-task remote Physiological measurement (\textbf{PhysMLE}). We begin with the problem formulation and challenge of multi-source synsemantic domain generalization in Subsection A. Subsequently, the framework of PhysMLE, including the overall structure of how PhysMLE can be integrated into a substitutable backbone network, the internal structure of the PhysMLE layer, and the router is provided in Subsection B, where we also verify the idea of learning unique low-rank parameters for each task. We then elaborate on the regularizations to maintain semantically spatio-temporally consistent and eliminate the cross-domain environmental variation in Subsection C. Lastly, subsection D introduces the optimization and inference processes. Important notations and descriptions are in Table \ref{t1}.

\begin{figure*}
\begin{center}
\includegraphics[scale=0.2]{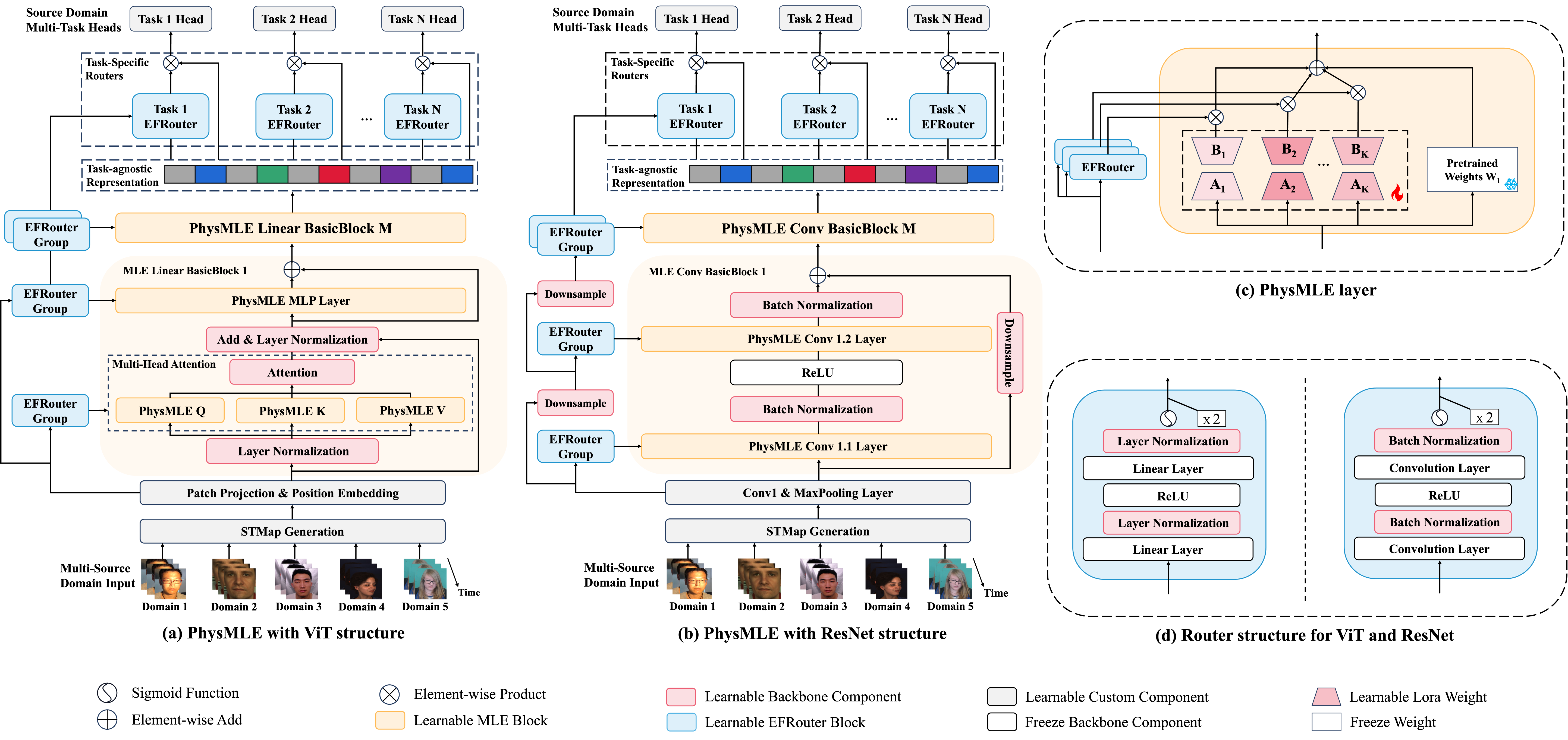}\\
\end{center}
\vspace{-3mm}
\caption{The overall architecture of proposed PhysMLE. Our method achieves flexibility over two types of backbone network structures (subfigure (a)(b)). Briefly, PhysMLE takes STMaps that are compressed from facial videos from multiple domains as input. The core feature learning layers in each basic block of the backbone are replaced by the PhysMLE layer (subfigure (c)). To adapt different backbone structures, the proposed EFRouter can also be instantiated as two types (subfigure (d)).}\label{f2}
\vspace{-3mm}
\end{figure*}

\subsection{Problem Formulation}
Here we define the notations and problem settings of our study. As shown in Figure \ref{f2}, suppose we have a batch of $B$ raw facial videos, which are from multiple datasets. We first compress videos into the spatial-temporal map (STMap) format \cite{niu2019rhythmnet}. Then we input STMaps $X = \{x_i\}_{i=1}^{B}, x_i \in \mathbb{R}^{L\times W\times C}$ into PhysMLE, which is formulated as $f(X;\theta)$. The $L, W, C$ is the length, width, and channel size of STMap, and $\theta$ is all parameters in the model. There are $N=4$ predicted vital signs of the subject in the target domain $Y = \{y_{hr}, y_{bvp}, y_{spo}, y_{rr}\}$ in this work, where $y_{hr}$ indicates HR, $y_{bvp}$ is the BVP, $y_{spo}$ and $y_{rr}$ corresponds to SpO2 and RR, respectively. 

In the field of remote physiological measurement, particularly in the context of multi-task video-based assessment in a cross-domain setting, one of the notable challenges stems from the domain shift. While the generalized principle of rPPG-based physiological measurements \cite{verkruysse2008remote} provides opportunities for leveraging data from diverse contexts, it introduces a challenge: the differing skin properties, lighting conditions, camera quality, and environmental factors lead to changes in color intensity captured by an RGB camera in different scenarios. Additionally, physiological signals related to HR, BVP, RR, and SpO2 often exhibit overlapping frequency components. For instance, HR and BVP share frequency bands, as both are influenced by cardiac activity. Thus, the cross-contamination between physiological signals adds another layer of complexity. Specifically, changes in one physiological parameter can affect or mask the variations in another parameter. For example, changes in SpO2 can impact the BVP signal, as oxygenation levels influence the light absorption properties of tissues \cite{mannheimer2007light}. Therefore, the goal of our model $f$ is to maximize the effective changes in color intensity while minimizing interference from other signals and domain variations simultaneously.

\subsection{PhysMLE}
\textbf{Mixture of Low-rank Expert} In this section, we introduce the concept of Mixture-of-Experts (MoE) \cite{shazeer2016outrageously}, proposing a unified \textbf{PhysMLE} basic block adaptive to the backbone network with an element-wise gate routing strategy. To learn the underlying relationships between different physiological signals, PhysMLE basic blocks are shared for the further estimation of each task.

In a typical MoE structure, multiple expert blocks are injected into one layer of the model. Although this method accommodates a greater number of parameters and overcomes the limitation of a singular set of parameters across all tasks, the parameter number to be tuned will increase exponentially. Recently, LoRA has demonstrated both its effectiveness and efficiency in fine-tuning, with its integral parameters being fine-tuned for all tasks. 

Thus, as shown in Figure \ref{f2}(c), the proposed PhysMLE basic block seamlessly integrates the advantages of both LoRA and MoE. Firstly, similar to LoRA for large language models (LLM) \cite{hu2021lora}, we reformulate the parameter update with the equation $\textbf{W} + \textbf{W}_\Delta = \textbf{W}+\textbf{BA}$, where $\textbf{W}, \textbf{W}_\Delta \in \mathbb{R}^{d_{in}\times d_{out}}$ denote the frozen weight matrix  and the weight matrix that is updated during fine-tuning, respectively. $\textbf{B} \in \mathbb{R}^{d_{in}\times r}$ and $\textbf{A} \in \mathbb{R}^{r\times d_{out}}$ are low-rank trainable matrix with $r\ll d_{in}/d_{out}$. Here, $d_{in}, d_{out}$ are input and output vector dimensions, $r$ is the rank. Particularly, when LoRA adapted to convolutional network \cite{zhong2023convolution}, $d_{in}, d_{out}$ are replaced by $C_{in}, C_{out}$, which present the number of input and output channels. In our PhysMLE, we introduce a set of low-rank experts ${E_i}^{K}_{i=1}$, each $E_i$ consists of a pair of $\textbf{BA}$, and both two types of LoRA were implemented for different backbone networks. 

Besides, being different from recent MoE-based LoRA frameworks \cite{zhong2023convolution,chen2023octavius,liu2023moelora}, PhysMLE introduces an Element-wise Feature Router (\textbf{EFRouter}) group mechanism (see Figure \ref{f2}(d)), which aims to leverage low-level physiological features to dynamically aggregate element-level feature outputted by each expert. The previous gate mechanism in MoE is mostly based on applying the softmax function over the linearly transformed high-level representation of experts \cite{liu2023moelora, hu2021lora}. However, given the strong association between physiological signals and facial low-level features (e.g., movement \cite{li2023learning}, expression \cite{casado2023depression}), the gate mechanism of previous MoE is difficult to be applied to remote multitasking physiological measurement. Thus, inspired by \cite{chang2023pepnet}, we designed the EFRouter groups $\{G_i\}_{i=1}^{K}$, where each $G_i$ corresponds to one expert $E_i$, to adaptively control the importance of low-level information, and used the hyperparameter $\gamma=2$ to further squash and expand the effective feature elements of the corresponding expert $E_i$. Each $G_i$ takes the low-level feature $s$ outputted by the first layer of the backbone network (i.e., Conv1 in ResNet \cite{he2016deep}, Patch Projection \& Position Embedding layers in ViT \cite{dosovitskiy2020image}) as the input. The $G_i$ consists of two neural layers, which are with one normalization layer, and one non-linear activation ($\operatorname{ReLU}$). After passing through the two layers, $\operatorname{Sigmoid}$ function was used to generate gated vectors. In all, the process of generating the output $s^{'}$ of each PhysMLE layer is customized as follows:

\begin{equation}
\label{eq1}
\begin{aligned}
s^{'} &= \textbf{W}s + \frac{\alpha}{r} \cdot \textbf{W}_{\Delta}s, \\
&= \textbf{W}s + \frac{\alpha}{r} \cdot \sum_{i=1}^{K}\gamma \cdot G_i(s) \cdot E_i(s), \\
&= \textbf{W}s + \frac{\alpha}{r} \cdot \sum_{i=1}^{K}\gamma  \cdot G_i(s) \cdot \textbf{B}_{i}\textbf{A}_{i}s.
\end{aligned}
\end{equation}

where $\alpha$ is the constant hyperparameter facilitating the tuning of rank $r$, and $\cdot$ is element-wise dot product. Note that, as the convolution layer has the stride of 2 in ResNet \cite{he2016deep}, we performed the downsampling layer of each block.

The primary objective of PhysMLE is to estimate $N=4$ physiological signals (i.e., HR, BVP, SpO2, and RR) simultaneously. After the STMaps, $X$ traverses through the multiple PhysMLE basic blocks and acquires the task-agnostic high-level feature $s^{'}$. Then, we inputted this feature into $N=4$ task-specific EFRouters. These routers are responsible for generating task-specific features $s^{''}=\{s^{''}_{hr}, s^{''}_{bvp}, s^{''}_{spo}, s^{''}_{rr}\}$, wherein each task head is dedicated to the regression of the corresponding physiological signal.

\textbf{Experts Orthogonalization} Furthermore, the pivotal consideration in resolving multi-task remote physiological measurement is to identify specific distinct features associated with each physiological signal. Thus, motivated by \cite{shah2023ziplora}, we noticed that directly adding the non-orthogonal columns to each other would lead to superimposition of their information about the individual physiological signal, making it challenging to disentangle features for each task. Being different from the multi-scale MoE \cite{zhong2023convolution}, we designed an Orthogonal Experts Merge (\textbf{OEM}) regularization over each expert's weights in the same layer to further minimize computational overhead, where the Spectral Restricted Isometry Property Regularization (SRIP) \cite{bansal2018can} was instantiated. For instance, given $M$ basic blocks in the backbone network and $T$ PhysMLE layers in each basic block, if there is $K=2$ expert weights $\mathbf{W_1}, \mathbf{W_2} \in \mathbb{R}^{d_{in}\times d_{out}}$ initialized in each PhysMLE layer, the $\mathcal{L}_{OEM}$ was given by:

\begin{equation}
\label{eq2}
\begin{aligned}
&u = (\mathbf{W_{1}}\mathbf{W_{2}}^{T} - (I- \text{{diag}}))\epsilon, \\
&v = (\mathbf{W_{1}}\mathbf{W_{2}}^{T} - (I- \text{{diag}}))u, \\
&\mathcal{L}_{OEM} = \frac{1}{M \times T}\sum_{i=1}^{M \times T}\frac{{\left\| v_i \right\|_2}} {{\left\| u_i \right\|_2 + 1e^{-12}}}.
\end{aligned}
\end{equation}

The $I \in \mathbb{R}^{d_{in}\times d_{in}}$ is the a matrix of ones, $\text{{diag}} \in \mathbb{R}^{d_{in}\times d_{in}}$ is an identity matrix, and $\epsilon \in \mathbb{R}^{d_{in}\times 1}$ is a random noise vector sampled from Gaussian distribution.

\begin{figure}
\begin{center}
\includegraphics[scale=0.53]{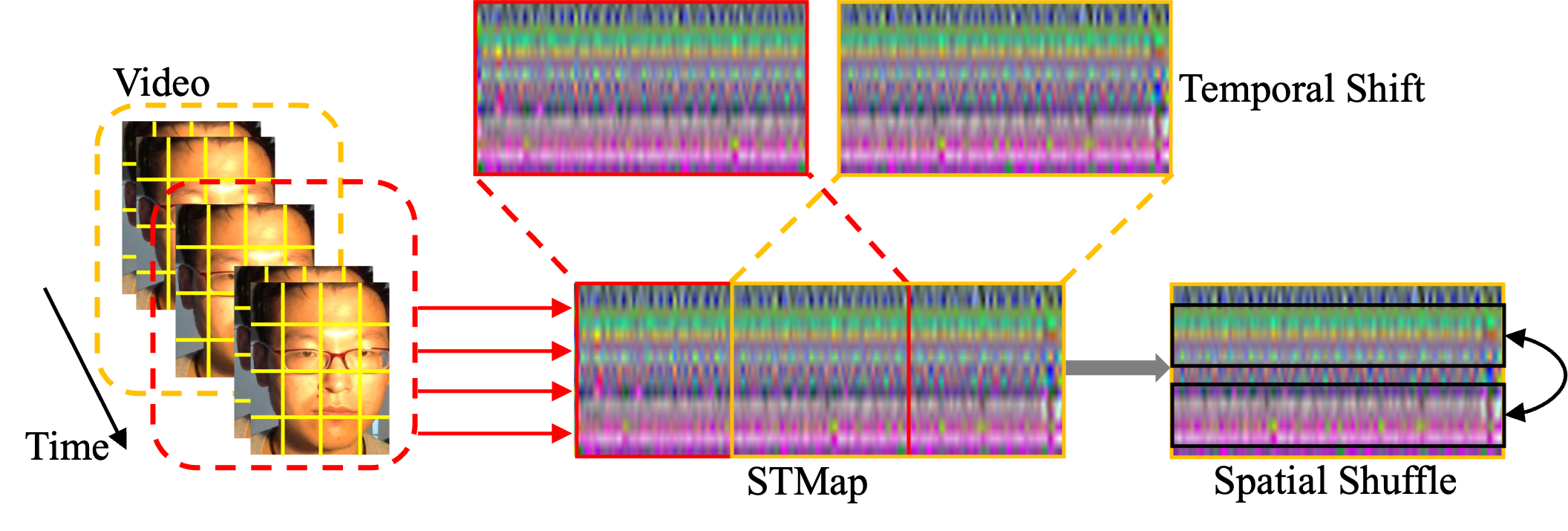}\\
\end{center}
\vspace{-6mm}
\caption{Illustration of how STMap generation and data augmentation performed. After the 256-frame facial video (highlighted with the red box) is aligned and compressed to an STMap, we shifted the sliding window to the next 50 frames to get the augmented STMap (highlighted with the orange box). Meanwhile, for the augmented STMap, the row-wise shuffle was applied.}\label{f3}
\vspace{-3mm}
\end{figure}

\subsection{Shared Semantic Feature Consistent}
Previous studies have highlighted the presence of analogous waveforms in short-term rPPG signals captured from different facial regions \cite{lam2015robust,tulyakov2016self,kumar2015distanceppg}. This observation serves as a crucial piece of prior knowledge in the realm of remote physiological measurement. While these analogous waveforms have been extensively utilized for heart rate measurements \cite{sun2024, sun2022contrast}, we propose that this knowledge can be leveraged to enhance the generalizability of multi-task physiological measurement, specifically, through regularization and data augmentation. 

First, inspired by \cite{chen2020simple}, we applied the spatio-temporal augmentation following our prior work \cite{lu2023neuron}. As shown in Figure \ref{f3}, after shuffling each row of STMap generated from RGB facial video \cite{niu2019rhythmnet} and sliding the time window with the size of 30 frames, we obtained the original input $X$ and augmented one $X^{'}$ corresponding to specific physiological states $Y$ and $Y^{'}$, respectively. Then, being different from previous attempts \cite{sun2024, sun2022contrast}, we maintained the spatial consistency over the aggregated features outputted by each layer of PhysMLE to encourage consistent predictions across different facial regions. Specifically, take the MLE Conv layer as an example, for the feature matrix $s \in \mathbb{R}^{B\times C_j\times W_j\times L_j}$ decoded by the j-th PhysMLE basic block, we regularized it with the Semantically Spatial Consistent loss as follows:

\begin{equation}
\label{eq3}
\mathcal{L}_{Spatial} = - \frac{1}{B \times C_j \times W_j}\sum_{i=1}^{B}\sum_{c=1}^{C_j}\sum_{w=1}^{W_j}\operatorname{sim}(s_{icw}, s_{icw^{*}}). 
\end{equation}

where $w^* \in random \{1,2,...,W_j\}$, and $\operatorname{sim}$ indicates the semantic similarity measurement, which is instantiated with cosine similarity. Then, we took the sum $\mathcal{L}_{Spatial}$ of all PhysMLE layers as the final $\mathcal{L}_{Spatial}$. Moreover, after getting estimated $\hat{Y}$ and $\hat{Y}^{'}$ based on $X$ and $X^{'}$, we applied a $\mathcal{L}_{Temporal}$ to penalize outputs, where $\hat{Y}^{'}$ was deviated from $\hat{Y}$ by more than the hyper-parameter $\delta$. This loss function aims to regularize the model following the temporal smoothness of physiological features within a short time interval.

\begin{equation}
\label{eq4}
\mathcal{L}_{Temporal} = \frac{1}{B}\sum_{i=1}^{B}\sum_{j=1}^{N} \left|\hat{Y}^{'}_{ij}  - \hat{Y}_{ij} \right|, where \ \left|\hat{Y}^{'}_{ij}-\hat{Y}_{ij}\right| \geq \delta.
\end{equation}

\subsection{Physiological Priors for Imbalance Target}
Referring to Table \ref{t0}, we can notice that only ground-truth HR is widely provided, particularly under our MSSDG setting, while other vital signs (i.e., HR, SpO2, and RR) are partially offered. To overcome the imbalance optimization goal when jointly training from multiple datasets, we leverage the prior knowledge from the physiology field to provide valid gradients from tasks that lack sufficient labels.

\textbf{Priors for SpO2} Lack of supervisable SpO2 labeling is present in large-scale multi-task remote physiological monitoring based on publicly available datasets. Under our setting, only PURE and VIPL-HR provided SpO2 labels. Particularly when conducting MSSDG protocol, there will be one dataset with SpO2 labels that can be leveraged to supervise the training of the SpO2 estimation head. Further, despite the existence of some correlation between SpO2 and HR, RR, and BVP, no direct causal relationship is there. Therefore, we proposed a novel data augmentation strategy based on prior knowledge of remote SpO2 measurement.

According to \cite{tarassenko2014non, cheng2024contactless}, the principle of remote SpO2 measurement with RGB camera relies on Eq. (\ref{eq5}). Where $AC_{RED}, AC_{BLUE}$ represent the standard deviations of the blue and red color channels, respectively, while $DC_{RED}, DC_{BLUE}$ represent the mean of the blue and red color channels. The coefficients $\mathbf{W_a}$ and $\mathbf{W_b}$ were determined by learning the best-fit linear correlation between the ratios of the red and blue channels and the ground-truth SpO2.

\begin{equation}
\label{eq5}
y_{spo} = \mathbf{W_a} - \mathbf{W_b}\frac{(AC_{RED}/DC_{RED})}{(AC_{BLUE}/DC_{BLUE})}.
\end{equation}

Back to our work, given the forward process of SpO2 estimation, we can argue that: (1) the ratio $\frac{AC_{RED}/DC_{RED}}{AC_{BLUE}/DC_{BLUE}}$ can be extracted by some of the experts in PhysMLE basic blocks, and it can be formulated in $s^{''}_{spo}$ after the SpO2-specific EFRouter; (2) through training, the $\mathbf{W_{spo}}$ of the SpO2 estimation head should be able to represent the $\mathbf{W_a}$ and $\mathbf{W_b}$, and reformulate Eq. (\ref{eq5}) to $y_{spo} = \mathbf{W_{spo}}s^{''}_{spo}$. Based on this, we constructed augmentation samples for SpO2 estimation through transformation in the ratio part. Specifically, given a $s^{''}_{spo}$ and its corresponding ground-truth $y_{spo}$, we defined a set of augmentation rate $\tau \in \mathrm{T}$, which is the set of transformations that change $s^{''}_{spo}$ with an arbitrary rate that is feasible under the Nyquist sampling theorem. We dot product $s^{''}_{spo}$ by $\tau$ to obtain a augmentation feature. Theoretically, the well-trained $\mathbf{W_{spo}}$ ought to map $\tau \cdot s^{''}_{spo}$ to the estimation that is also $\tau$ times to $y_{spo}$. Such augmentation strategy effectively changes the underlying color-channel ratios, thus creating different samples from the same individual but with different SpO2. In practice, we limit the augmented SpO2 range to be within [80\%, 100\%], ensuring the augmented SpO2 is not out of the normal range \cite{suprayitno2019measurement}. Given this, we designed the $\mathcal{L}_{ASp}$ as follows:

\begin{equation}
\label{eq6}
\mathcal{L}_{ASp} = \frac{1}{B}\sum^{B}\sum^{\mathrm{T}}_{\tau=1}  \left|\mathbf{W_{spo}}\tau \cdot s^{''}_{spo} - \tau \cdot y_{spo} \right|.
\end{equation}

\begin{figure}
\begin{center}
\includegraphics[scale=0.44]{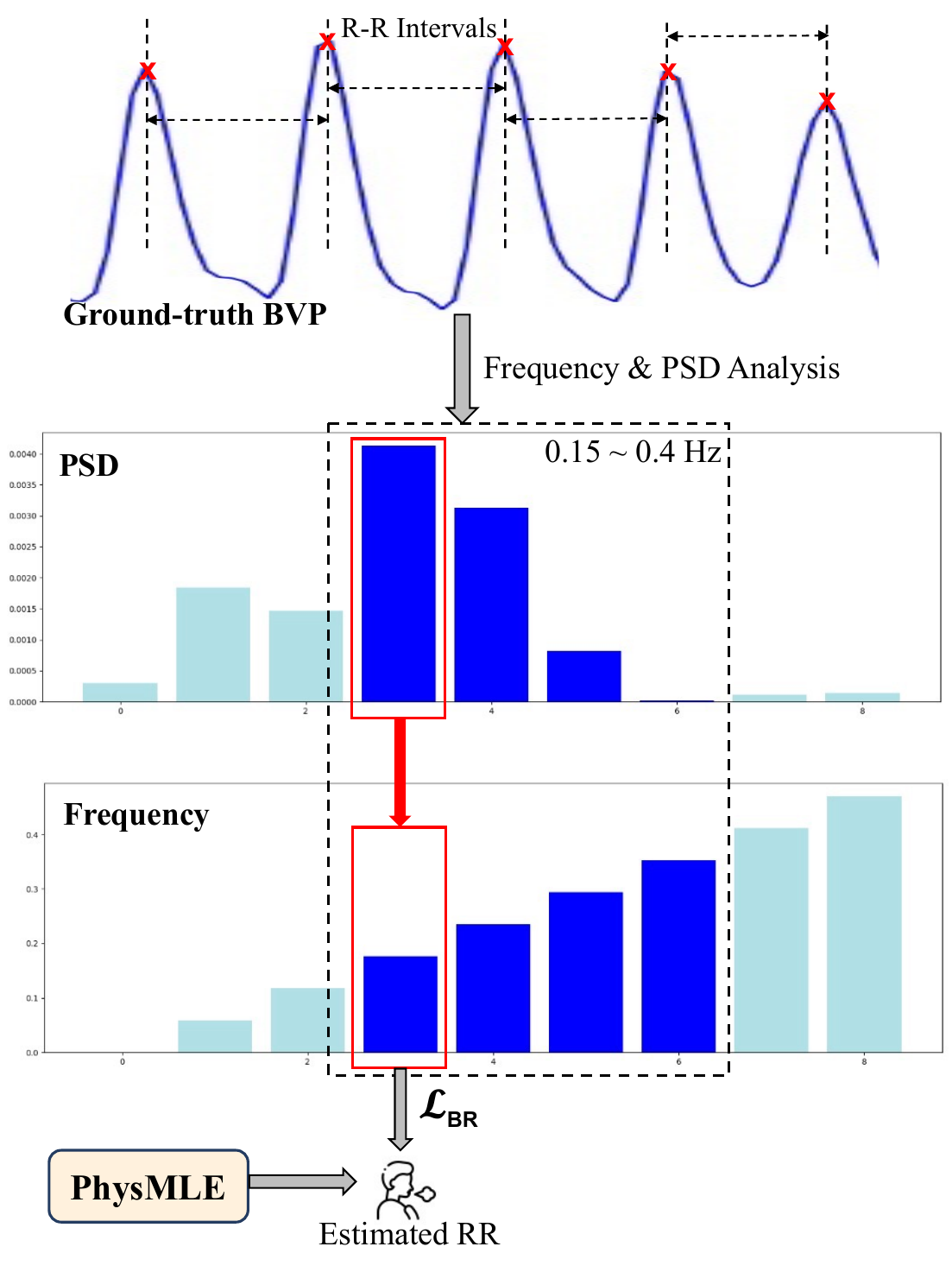}\\
\end{center}
\vspace{-3mm}
\caption{Example elaboration of $\mathcal{L}_{BR}$ calculation. For those samples with ground-truth BVP signals but without RR, we first detect R-R intervals and conduct frequency and PSD analysis to extract $y^{bvp}_{rr}$ from the high-frequency part. Then, we use $y^{bvp}_{rr}$ to supervise the learning of RR estimation according to Eq. (\ref{eq7}).}\label{f6}
\vspace{-3mm}
\end{figure}

\textbf{Priors for RR} As the respiratory process affects the autonomic nervous system, it further leads to periodic changes in the heart rate, which is defined as respiratory sinus arrhythmia (RSA) \cite{stauss2003heart}. Meanwhile, it is believed that the High Frequency (HF) component in HRV (i.e., 0.15-0.4Hz) reflects the parasympathetic influence on the heart and is significantly associated with RSA \cite{berntson1997heart}. Given this, previous works \cite{li2018obf,niu2020CVD} assessed the RR by using the peak location of the HF component from the power spectrum densities (PSD) of the R-R intervals. Similar to $\mathcal{L}_{HB}$, we propose the $\mathcal{L}_{BR}$ to augment the training of unsupervised RR samples with labels calculated from ground-truth BVP. However, most of the current rPPG works do not continuously output long-period BVP signals during training due to computational efficiency and instantaneous HR estimation, while the precise HRV calculation usually requires over 300s \cite{bernardes2022reliable}. Thus, refer to Figure \ref{f6}, different from $\mathcal{L}_{HB}$, we softly constraint the difference between the $\hat{y}_{rr}$ and the extracted $y_{rr}^{bvp}$ from ground-truth BVP should be within $\delta$. 

\begin{equation}
\label{eq7}
\mathcal{L}_{BR} = \frac{1}{B}\sum^{B} \left|\hat{y}_{rr}  - y_{rr}^{bvp} \right|, where \ \left|\hat{y}_{rr} -y_{rr}^{bvp}\right| \geq \delta.
\end{equation}

\textbf{Priors for HR and BVP} Specifically, the basic principle of rPPG is that when pulse waves triggered by cardiac activity reach the surface of the skin, they cause small color changes that can be captured by an RGB camera \cite{verkruysse2008remote}, and BVP signals, which reflect the shape and intensity of the pulse waves, contain rich information about cardiac activity. Thus, HR can be directly calculated by the number of peaks in the BVP signal and can be calculated by peak detection or spectral analysis. Therefore, given the HR label, when there is no BVP label provided for some samples, we reinforce the learning of the BVP head by constraining the HR calculated from the estimated BVP signal to be consistent with the ground-truth HR. For the design of the optimization objective $\mathcal{L}_{HB}$, we deliberately used the peak detection to compute HR from estimated BVP, which is beneficial to the computational efficiency on the one hand and on the other hand, such sensitivity-to-noise method can force the network to inference the BVP signals that are as free of noise as possible during the training process. 

\begin{equation}
\label{eq8}
\mathcal{L}_{HB} = \frac{1}{B}\sum^{B} \left|\hat{y}_{hr}  - FindPeak(\hat{y}_{bvp}) \right|.
\end{equation}

\subsection{Optimization and Inference}
In this section, we provide a detailed explanation of the optimization process of PhysMLE. To enhance clarity, we present the entire procedure in Algorithm 1. 

To facilitate the estimation of HR, SpO2, and RR, we formulate three loss functions, denoted as $\mathcal{L}_{HR}$, $\mathcal{L}_{SpO2}$, and $\mathcal{L}_{RR}$, respectively. These loss functions employed the L1 loss to predict the mean values of the respective signal distributions. To learn the BVP signal feature, we utilized the negative Pearson's correlation coefficient as the basis for the $\mathcal{L}_{BVP}$. During the joint training process, we incorporated an adaptation factor $\lambda$ \cite{lu2023neuron} to suppress meaningless regularizations during the early iterations. By combining all the individual loss functions, we formulated the overall loss function with trade-off parameters $p_i$ for each specific regularization as:

\begin{equation}
\label{eq9}
\begin{aligned}
\mathcal{L}(X, Y) &= \mathcal{L}_{BVP} + \lambda(p_1\mathcal{L}_{HR} + p_2\mathcal{L}_{SpO2} + p_3\mathcal{L}_{RR} \\
&+ p_4\mathcal{L}_{OEM} + p_5\mathcal{L}_{Spatial} + p_6\mathcal{L}_{Temporal} \\
&+ p_7\mathcal{L}_{HB} + p_8\mathcal{L}_{BR} + p_9\mathcal{L}_{ASp}).
\end{aligned}
\end{equation}

\begin{algorithm}[h]
  \caption{Train Process of PhysMLE} 
  \begin{algorithmic}[1]
  \Require
  \Statex Initialize backbone network;
  \Statex Replace basic blocks with PhysMLE with $\alpha, r, K$;
  \Statex freeze part of parameters $\theta_f$ and activate $\theta_t$;
      \Statex Generated STMaps $X$ from videos;
      \Statex The ground-truth signals $Y$; 
      \Statex The number of iteration $N_{iter}$; 
      \Statex Hyper-parameters $\tau, \delta, \gamma, p$.
      \Ensure
      \For{ i = 1 to $N_{iter}$}
      \State Conduct data augmentation $X^{'}$ from to $X$ as shown in Figure \ref{f3};
      \State Encode $X, X^{'}$ to $s_0$ by the first non-PhysMLE layer;
      \State Forward $s_0$ in $M$ PhysMLE basic blocks;   
      \State Compute $\mathcal{L}_{OEM}$ based on $K$ expert weights in each PhysMLE layers by Eq. (\ref{eq2});
      \State Compute $\mathcal{L}_{Spatial}$ based on the outputted feature of each PhysMLE basic block by Eq. (\ref{eq3});
      \State Input the task-agnostic high-level feature $s^{'}$ into each task head and corresponding EFRouter to predict $\hat{Y},\hat{Y}^{'}$;
      \State Compute $\mathcal{L}_{Temporal}$ according to Eq.(\ref{eq4});
      \State Compute $\mathcal{L}_{HB}, \mathcal{L}_{RB}, \mathcal{L}_{ASp}$ according to Eq.(\ref{eq6})(\ref{eq7})(\ref{eq8}) respectively;
      \State Jointly update $\theta_t$ with the gradient from Eq.(\ref{eq9});
      \EndFor.
  \end{algorithmic}
\end{algorithm}

\begin{figure}
\begin{center}
\includegraphics[scale=0.25]{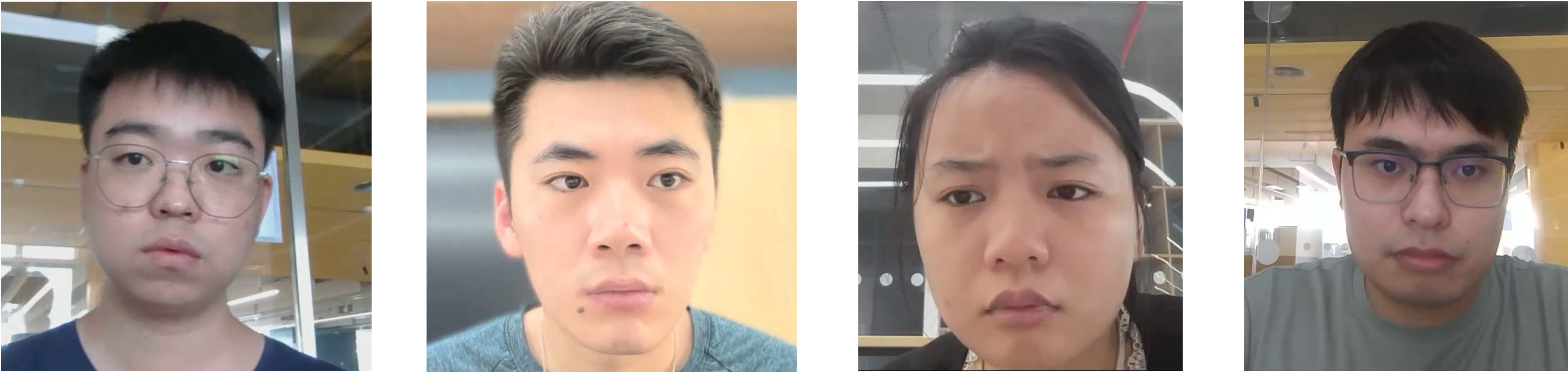}\\
\end{center}
\vspace{-3mm}
\caption{Sample video frames captured by different webcams in HCW.}\label{f4}
\vspace{-3mm}
\end{figure}

\begin{figure}
\begin{center}
\includegraphics[scale=0.32]{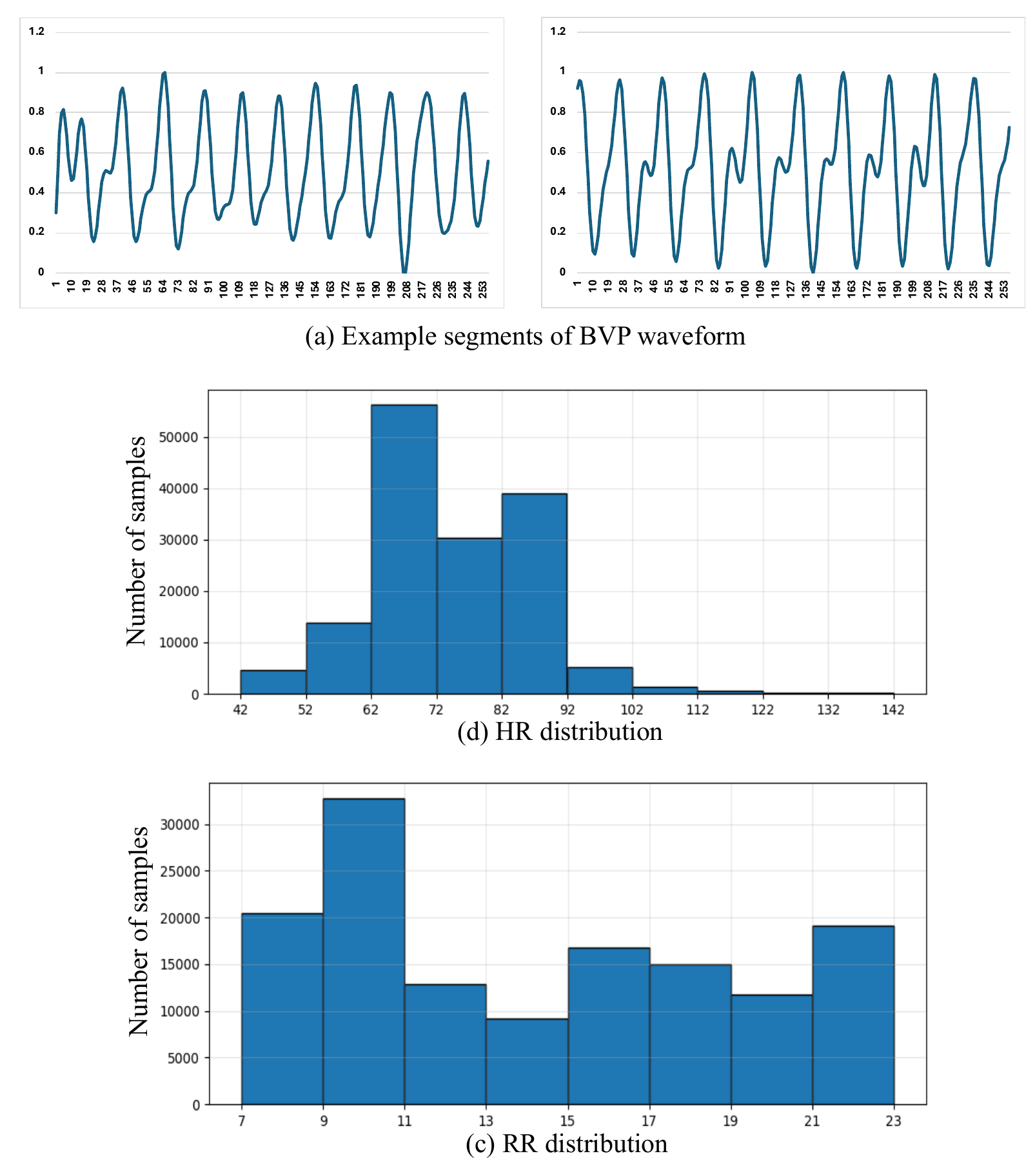}\\
\end{center}
\vspace{-6mm}
\caption{BVP waveform examples and statistical distribution of HR and RR in HCW.}\label{f5}
\vspace{-3mm}

\end{figure}

\begin{table*}[t]
\setlength{\tabcolsep}{1mm}
\centering
\scriptsize
\caption{Comparison Experiment Results on MSSDG Protocol I.}
\begin{tabular}{cccccccccccccc}
\toprule 
                          & & \multicolumn{6}{c}{\textbf{PURE}}             & \multicolumn{6}{c}{\textbf{VIPL-HR}}\\
\cmidrule(lr){3-8} \cmidrule(lr){9-14} 
 &      & \multicolumn{3}{c}{\textbf{HR}}    & \multicolumn{3}{c}{\textbf{SpO2}}   &     \multicolumn{3}{c}{\textbf{HR}}    & \multicolumn{3}{c}{\textbf{SpO2}}     \\
                    \textbf{Method Type}             & \textbf{Method} & \textbf{MAE↓}  & \textbf{RMSE↓} & \textbf{\quad p↑ \quad}    & \textbf{MAE↓}  & \textbf{RMSE↓} & \textbf{\quad p↑ \quad}    & \textbf{MAE↓}  & \textbf{RMSE↓} & \textbf{\quad p↑ \quad}    & \textbf{MAE↓}  & \textbf{RMSE↓}  & \textbf{\quad p↑ \quad}    \\
                    \midrule 
\textbf{Single-task} & \textbf{GREEN \cite{verkruysse2008remote} }       & 10.32 & 14.27 & 0.52 & —   & —   & —   & 12.18 & 18.23 & 0.25 & —    & —    & —   \\
\textbf{Traditional}& \textbf{CHROM \cite{de2013robust}  }      & 9.79  & 12.76 & 0.37 & —   & —   & —   & 11.44 & 16.97 & 0.28 & —    & —    & —   \\
 & \textbf{POS \cite{wang2016algorithmic} }         & 9.82  & 13.44 & 0.34 & —   & —   & —   & 14.59 & 21.26 & 0.19 & —    & —    & —   \\
                        & \textbf{ARM-SpO2 \cite{tarassenko2014non}}           & —    & —    & —   & 3.25 & 3.51 & 0.20  & —    & —    & —   & 18.11  & 20.23 & 0.01 \\
                        \midrule 
\textbf{Single-task}        & \textbf{RhythmNet \cite{niu2019rhythmnet} }   & 7.89  & 10.66 & 0.62 & —   & —   & —   & 10.97 & 14.86 & 0.31 & —    & —    & —   \\
\textbf{Deep} & \textbf{BVPNet \cite{das2021bvpnet} }     & 7.44  & 10.51 & 0.64 & —   & —   & —   & 10.62 & 14.1  & 0.32 & —    & —    & —   \\
                        & \textbf{Dual-GAN \cite{lu2021dual} }    & 7.65  & 10.86 & 0.63 & —   & —   & —   & 10.77 & 14.81 & 0.31 & —    & —    & —   \\
                        & \textbf{PhysFormer++ \cite{yu2023physformer++}} & 7.16  & 10.14 & 0.65 & —   & —   & —   & 9.91  & 14.05 & 0.37 & —    & —    & —   \\
                        & \textbf{EfficientPhys \cite{liu2023efficientphys}} & 8.21  & 11.30 & 0.57 & —   & —   & —   & 11.16  & 14.95 & 0.30 & —    & —    & —   \\
                        & \textbf{rSPO \cite{akamatsu2023blood}}         & —    & —    & —   & 3.41 & 3.60  & 0.18 & —    & —    & —   & 14.36 & 15.54 & 0.02 \\
                        \midrule 
\textbf{Multi-task}           & \textbf{AD-R \cite{ganin2015unsupervised}}           & 7.52  & 10.79 & 0.6  & 3.15 & 4.26 & 0.19 & 9.81  & 13.99 & 0.37 & 14.78 & 17.98 & 0.04 \\
\textbf{DG}  & \textbf{VREx-R \cite{krueger2021out}}     & 8.00     & 12.87 & 0.66 & 2.76 & 3.42 & 0.20  & 10.79 & 15.13 & 0.32 & 14.83 & 18.40  & 0.03 \\
                        & \textbf{GroupDRO-R \cite{parascandolo2020learning}}     & 8.98  & 10.52 & 0.62 & 3.20  & 4.22 & 0.19 & 15.65 & 17.8  & 0.22 & 16.30  & 21.48 & 0.01 \\
                        & \textbf{NEST-R \cite{lu2023neuron}}        & 6.57  & 10.03 & 0.67 & 3.57 & 4.17 & 0.19 & 10.44 & 14.69 & 0.33 & 15.61 & 18.78 & 0.03 \\
                        & \textbf{HSRD-R \cite{wang2023hierarchical} }       & 5.91  & 8.39  & 0.71 & 2.80  & 3.98 & 0.20  & 10.04 & 14.33 & 0.36 & 14.83 & 19.41 & 0.06 \\
                        & \textbf{AD-T \cite{ganin2015unsupervised} }          & 8.52  & 10.36 & 0.54 & 2.01 & 2.34 & 0.20  & 11.10  & 15.35 & 0.26 & 11.34 & 12.00    & 0.09 \\
                        & \textbf{VREx-T \cite{krueger2021out}}         & 9.29  & 11.22 & 0.48 & 2.48 & 2.99 & 0.19 & 11.72 & 16.11 & 0.26 & 14.68 & 15.12 & 0.03 \\
                        & \textbf{GroupDRO-T \cite{parascandolo2020learning} }    & 9.46  & 11.36 & 0.49 & 2.71 & 2.74 & 0.18 & 10.74 & 14.78 & 0.29 & 13.45 & 13.92 & 0.05 \\
                        & \textbf{NEST-T \cite{lu2023neuron}}         & 7.52  & 9.66  & 0.55 & 2.13 & 2.25 & 0.19 & 10.62 & 14.23 & 0.30  & 13.28 & 13.12 & 0.06 \\
                        & \textbf{HSRD-T \cite{wang2023hierarchical}}         & 7.31  & 9.25  & 0.58 & 2.26 & 2.39 & 0.19 & 10.36 & 14.19 & 0.33 & 12.25 & 12.66 & 0.08 \\
                        \midrule 
                       \textbf{Ours} & \textbf{Base-R \cite{he2016deep} }   & 8.74 & 10.11 & 0.56 & 3.93 & 4.60  & 0.18 & 9.96  & 14.24 & 0.36 & 19.05 & 19.25 & 0.01 \\
                        & \textbf{Base-T \cite{wu2020visual} }        & 8.81  & 10.21 & 0.55 & 2.45 & 2.85 & 0.19 & 11.36 & 15.40  & 0.25 & 14.61 & 15.76 & 0.03 \\
                       & \textbf{PhysMLE-R }    & \textbf{5.43}  & \textbf{7.88}  & \textbf{0.77} & 2.19 & 3.23 & 0.22 & \textbf{7.53}  & \textbf{10.29} & \textbf{0.62} & 10.94 & 14.81 & 0.09 \\
                        & \textbf{PhysMLE-T}     & 6.62  & 8.72  & 0.67 & \textbf{1.86} & \textbf{2.34} & \textbf{0.25} & 9.59  & 12.74 & 0.37 & \textbf{9.88}  & \textbf{10.77} & \textbf{0.10}\\
\bottomrule 
\end{tabular}
\textnormal{\\Notes: In this and following tables, the \textbf{bold} texts show the best result within each column.}
\label{t2}
\end{table*}

\begin{table*}[t]
\setlength{\tabcolsep}{1mm}
\centering
\scriptsize
\caption{Comparison Experiment Results on MSSDG Protocol I.}
\begin{tabular}{cccccccccccccc}
\toprule 
                          & & \multicolumn{6}{c}{\textbf{V4V}}             & \multicolumn{6}{c}{\textbf{HCW}}\\
\cmidrule(lr){3-8} \cmidrule(lr){9-14} 
 &      & \multicolumn{3}{c}{\textbf{HR}}    & \multicolumn{3}{c}{\textbf{RR}}   &     \multicolumn{3}{c}{\textbf{HR}}    & \multicolumn{3}{c}{\textbf{RR}}     \\
                    \textbf{Method Type}             & \textbf{Method} & \textbf{MAE↓}  & \textbf{RMSE↓} & \textbf{\quad p↑ \quad}    & \textbf{MAE↓}  & \textbf{RMSE↓} & \textbf{\quad p↑ \quad}    & \textbf{MAE↓}  & \textbf{RMSE↓} & \textbf{\quad p↑ \quad}    & \textbf{MAE↓}  & \textbf{RMSE↓}  & \textbf{\quad p↑ \quad}    \\
                    \midrule 
\textbf{Single-task} & \textbf{GREEN \cite{verkruysse2008remote} }       & 15.64         & 21.43          & 0.06 & —   & —   & —   & 14.36 & 20.01 & 0.04 & —    & —    & —   \\
\textbf{Traditional}& \textbf{CHROM \cite{de2013robust}  }      & 14.92         & 19.22          & 0.08 & —   & —   & —   & 14.66 & 19.81 & 0.05 & —    & —    & —   \\
 & \textbf{POS \cite{wang2016algorithmic} }         & 12.67         & 18.11          & 0.06  & —   & —   & —   & 12.37 & 17.61 & 0.11 & —    & —    & —   \\
                        & \textbf{ARM-RR \cite{tarassenko2014non}}           & —    & —    & —   & 6.33 & 6.90 & 0.01  & —    & —    & —   & 8.02  & 8.66 & 0.01 \\
                        \midrule 
\textbf{Single-task}        & \textbf{RhythmNet \cite{niu2019rhythmnet} }   & 11.31 & 15.37 & 0.30   & —   & —   & —       & 10.10 & 11.73 & 0.35   & —    & —    & — \\
\textbf{Deep} & \textbf{BVPNet \cite{das2021bvpnet} }     & 11.16 & 15.10 & 0.31   & —   & —   & —       & 9.52  & 12.01 & 0.57   & —    & —    & —   \\
                        & \textbf{Dual-GAN \cite{lu2021dual} }    & 10.02 & 14.23 & 0.32   & —   & —   & —       & 9.44  & 11.61 & 0.58   & —    & —    & —   \\
                        & \textbf{PhysFormer++ \cite{yu2023physformer++}} & 9.53  & 12.96 & 0.37   & —   & —   & —       & 8.12  & 10.05 & 0.51   & —    & —    & —   \\
                        & \textbf{EfficientPhys \cite{liu2023efficientphys}} & 11.87 & 16.02 & 0.27   & —   & —   & —       & 10.02 & 11.79 & 0.34   & —    & —    & —   \\
                        \midrule 
\textbf{Multi-task} & \textbf{MTTS-CAN \cite{liu2020multi} }     & 12.06 & 15.01 & 0.11   & 6.44 & 7.80 & 0.01     & 13.38 & 15.21 & 0.21   & 9.33  & 15.86 & 0.01   \\
\textbf{Deep} & \textbf{BigSmall \cite{narayanswamy2024bigsmall}}    & 11.88 & 14.60 & 0.29   & 6.36 & 7.68 & 0.02     & 12.05 & 14.11 & 0.25   & 8.42  & 14.30 & 0.01   \\
                        \midrule
\textbf{Multi-task}           & \textbf{AD-R \cite{ganin2015unsupervised}}           & 10.91 & 13.72 & 0.31   & 3.06 & 3.83 & 0.14     & 9.86 & 11.14 & 0.41   & 7.54  & 13.84 & 0.04 \\
\textbf{DG}  & \textbf{VREx-R \cite{krueger2021out}}     & 10.96 & 13.78 & 0.31   & 3.23 & 4.23 & 0.13     & 8.23 & 10.93 & 0.49   & 7.44  & 12.56 & 0.06 \\
                        & \textbf{GroupDRO-R \cite{parascandolo2020learning}}     & 10.58 & 12.92 & 0.32   & 3.98 & 4.47 & 0.12     & 9.11  & 11.39 & 0.44   & 8.28  & 13.17 & 0.03 \\
                        & \textbf{NEST-R \cite{lu2023neuron}}        & 9.62  & 11.05 & 0.38   & 3.46 & 4.50 & 0.12     & 9.76 & 11.35 & 0.41   & 7.31  & 12.59 & 0.04 \\
                        & \textbf{HSRD-R \cite{wang2023hierarchical} }       & 9.18  & 10.83 & 0.42   & 3.11 & 4.01 & 0.13     & 9.35  & 10.02 & 0.47   & 7.28  & 12.92 & 0.04 \\
                        & \textbf{AD-T \cite{ganin2015unsupervised} }          & 10.95 & 13.47 & 0.31   & 5.49 & 6.04 & 0.05     & 8.75  & 10.71 & 0.46   & 8.25  & 9.95  & 0.03 \\
                        & \textbf{VREx-T \cite{krueger2021out}}         & 11.21 & 13.61 & 0.25   & 6.68 & 8.22 & 0.02     & 8.41  & 10.32 & 0.49   & 10.43 & 14.68 & 0.02 \\
                        & \textbf{GroupDRO-T \cite{parascandolo2020learning} }    & 11.17 & 13.57 & 0.30   & 5.59 & 6.10 & 0.05     & 9.10  & 11.18 & 0.39   & 8.05  & 11.92 & 0.03 \\
                        & \textbf{NEST-T \cite{lu2023neuron}}         & 10.42 & 12.79 & 0.31   & 4.75 & 5.29 & 0.07     & 8.39  & 10.30 & 0.50   & 9.23  & 10.42 & 0.03 \\
                        & \textbf{HSRD-T \cite{wang2023hierarchical}}         & 10.32 & 13.40 & 0.33   & 4.98 & 5.20 & 0.07     & 8.30  & 10.27 & 0.49   & 8.44  & 10.21 & 0.03 \\
                        \midrule 
                       \textbf{Ours} & \textbf{Base-R \cite{he2016deep} }   & 11.64 & 14.05 & 0.28   & 4.15 & 4.87 & 0.08     & 10.18 & 11.30 & 0.36   & 8.22  & 12.25 & 0.04 \\
                        & \textbf{Base-T \cite{wu2020visual} }        & 11.30 & 14.04 & 0.26   & 4.81 & 6.32 & 0.07    & 10.24 & 12.96 & 0.35   & 9.74  & 12.94 & 0.05 \\
                       & \textbf{PhysMLE-R }    & \textbf{8.12}  & \textbf{10.56} & \textbf{0.51}   & \textbf{3.02} & \textbf{3.66} & \textbf{0.14}     & \textbf{7.13}  & 10.20 & \textbf{0.61}   & \textbf{5.90}  & \textbf{8.84}  & \textbf{0.10} \\
                        & \textbf{PhysMLE-T}     & 8.96  & 10.90 & 0.50   & 3.41 & 4.91 & 0.12     & 7.34  & \textbf{9.46}  & 0.60   & 6.55  & 9.48  & 0.06\\
\bottomrule 
\end{tabular}
\label{t3}
\end{table*}

\section{Experiment}
 
\subsection{Dataset}
\subsubsection{Our Collected Dataset}
Previous public rPPG datasets have played a significant role in advancing the field of remote physiological measurement. However, these datasets have certain limitations that restrict their application in real-world scenarios. One key limitation is the lack of consideration for the influence of human states on physiological signs. As summarized in Table \ref{t0}, many existing datasets ignore the dynamic nature of human physiological signs under varying cognitive or emotional states \cite{wang2020cognitive,basu2016effects}. Although a few datasets satisfied the demand of special human states (e.g., the stress in UBFC-Phys \cite{sabour2021ubfc}, emotions in MAHNOB-HCI \cite{soleymani2011multimodal}), they were recorded using specialist imaging devices. This restricts the generalizability of the models trained on these datasets to real-world situations where a wider range of webcam types are commonly involved. Therefore, we built a dataset collected in the real workspace, named \textbf{HCW}, to promote the remote physiological measurement under the high cognitive workload scenario. In total, 41 participants with 27 males and 14 females (aged between 22 and 31) engaged in the collection of HCW, where subjects were induced to the high cognitive workload status.

Specifically, as previous datasets already cover diverse environmental illumination and subject pose, we designed our data to have variations in cognitive workload levels. The data collection was conducted in a quite room with stable lighting conditions. In order to replicate daily uncontrollable application scenarios, we first asked subjects to sit casually as they do in their own place, then encouraged them to perform daily work activities (e.g., literature review and paper reading) with their own laptop. Then, during the 100-minute observation, we assessed their cognitive workload using the NASA-TLX with 25-minute constant intervals. Whenever they reported a NASA-TLX over 70 \cite{wang2020cognitive}, we deployed physiological signal sensors and opened the in-front webcam on their laptop to record the video of next 25 minutes at 30 fps. Some examples captured by different webcams can be found in Figure \ref{f4}. The integrated platform PhysioLAB was utilized for simultaneous physiological signal collection. The BVP signals were recorded with 1000Hz. Golden HR and RR were calculated from ECG and respiration signals collected by the SMD electrocardiograph and respiratory belt transducers of PhysioLAB.

Then, we performed noise removal and downsampled the original BVP signals to 30HZ to match the frame rate of the videos. For the purpose of enhancing the accessibility and usability of our dataset, we preprocessed the raw data and converted it into a MAT file format, which is compatible with both Matlab and Python. Some BVP examples and distributions of HR and RR are in Figure \ref{f5}. The videos were filmed with a resolution of $2160\times 1080$ pixels. Using the face detection and alignment approaches in \cite{niu2019rhythmnet}, each frame of aligned facial videos is with the size of $128\times 128$. The experiment was approved by the Hong
Kong University of Science and Technology’s Human and Artefacts Research Ethics Committee (protocol number: HREP-2023-0159), and We will share both the raw data (i.e., original video and raw physiological signals) and processed data with the research community.

\subsubsection{Other Datasets}
In addition to HCW, we carefully selected five other datasets to evaluate the performance of our PhysMLE within the context of large-scale MSSDG protocol. The chosen datasets encompassed a range of motion, camera, and lighting conditions, presenting a comprehensive evaluation scenario.
\begin{itemize}
    \item UBFC-rPPG \cite{bobbia2019unsupervised} consists of 42 facial videos captured under both sunlight and indoor illumination conditions. Ground-truth BVP signals and HR values in this dataset were collected using CMS50E, providing reliable reference measurements.
    \item BUAA \cite{xi2020image} aims to evaluate performance under different illumination conditions. Only data with illumination levels of 10 lux or greater are included, as underexposed images require specialized algorithms beyond the scope of this study.
    \item PURE \cite{stricker2014non} comprises 60 RGB videos featuring 10 subjects engaged in six different activities. To align the BVP signals with the videos, the signals are down-sampled from 60 to 30 fps using cubic spline interpolation.
    \item VIPL-HR \cite{niu2019rhythmnet} offers nine distinct scenarios captured using three RGB cameras, reflecting various illumination conditions and movement levels. To address the issue of unstable video frame rates, we normalize the STMap to 30 fps using cubic spline interpolation, incorporating insights from prior works \cite{lu2021dual,yu2023physformer++}.
    \item V4V \cite{revanur2021first} is designed to capture significant changes in physiological indicators. It includes data from ten tasks.
\end{itemize}

Notably, as V4V only had HR values but without BVP signal labels and the BVP signals and videos in VIPL-HR are not temporally aligned, the $\mathcal{L}_{BVP}$ are inapplicable for these two datasets.

\subsection{Implementation Details}

\subsubsection{Training Settings}
The whole work was implemented by the Pytorch framework. For the data preparation, we followed the procedures of STMap generation in \cite{lu2023neuron}. The STMap of each dataset was sampled every 10 steps with a time window of 256. To comprehensively evaluate the effectiveness of PhysMLE adapted to different backbone networks, we created two versions of PhysMLE: \textbf{PhysMLE-R} adopted from ResNet-18 \cite{he2016deep} and \textbf{PhysMLE-T} using Vision Transformer pretrained with ImageNet \cite{wu2020visual}, and their core feature extraction layers were replaced by the PhysMLE layer. For PhysMLE-ResNet, the original STMap $\mathbb{R}^{25\times 256\times 3}$ was resized to $\mathbb{R}^{64\times 256\times 3}$, and $\mathbb{R}^{224\times 224\times 3}$ for PhysMLE-T. For HR, SpO2, and RR regression, one full connection layer was initialized for each task. At the same time, the BVP head consists of four blocks, which followed the design in \cite{lu2021dual}. The whole experiments were conducted on an RTX A6000. The batch size $B$ and iteration number $N_{iter}$ were 60 and 20000. Hyper-parameter $\tau$ and $\delta$ were set to 0.1 and 5. The number of experts $K$ was 3. The other two hyper-parameters $\alpha$ and  $r$ controlling the size of LoRA weights were selected from [8, 16, 32] depending on performance. Adam optimizer was used for training, with a learning rate of 0.00001.

\subsubsection{Evaluation Settings}
In testing, following existing works \cite{lu2023neuron, wang2023hierarchical}, we used mean absolute error (MAE), root mean square error (RMSE), and Pearson’s correlation coefficient (p) to assess estimated physiological indicators. Furthermore, for the HRV measurement, we used normalized low frequency (LFnu), normalized high frequency (HFnu), and LF/HF (i.e., the ratio between low-frequency power and high-frequency power).

The MSSDG protocol was designed into two parts: (1) \textbf{Protocol I}, this protocol is implemented in a cross-domain evaluation scenario by dividing six datasets into two groups: five datasets for training (source domain) and then testing on the other dataset (target domain); (2) \textbf{Protocol II}, this protocol is implemented in an intra-dataset multi-task evaluation scenario based on large challengable datasets. Specifically, two datasets (i.e., VIPL-HR and HCW) with labels not limited to BVP and HR were used as training and testing sets simultaneously. It is worth noting that, to avoid unfair comparison, we removed the action units prediction head of BigSmall \cite{narayanswamy2024bigsmall} and the GRU mechanism from Rhythmnet \cite{niu2019rhythmnet}.

\begin{table*}[]
\setlength{\tabcolsep}{0.6mm}
\caption{HRV and HR estimation results on MSSDG protocol I.}
\label{t4}
\centering
\scriptsize
\begin{tabular}{clcccccccccccc}
\toprule
 &   & \multicolumn{3}{c}{\textbf{LFnu}}   & \multicolumn{3}{c}{\textbf{HFnu}}   & \multicolumn{3}{c}{\textbf{LF/HF}} & \multicolumn{3}{c}{\textbf{HR-(bpm)}}     \\
\cmidrule(lr){3-5} \cmidrule(lr){6-8} 
\cmidrule(lr){9-11} \cmidrule(lr){12-14}        
{\textbf{Target}}&{\textbf{Method}}  & \textbf{MAE↓}    & \textbf{RMSE↓}   & \textbf{\quad p↑ \quad}      & \textbf{MAE↓}    & \textbf{RMSE↓}   & \textbf{\quad p↑ \quad}     & \textbf{MAE↓}    & \textbf{RMSE↓}   & \textbf{\quad p↑ \quad}     & \textbf{MAE↓}    & \textbf{RMSE↓}   & \textbf{\quad p↑ \quad}     \\
\midrule 
\textbf{UBFC} & 
\textbf{GREEN~\cite{verkruysse2008remote}}    & 0.236 & 0.284 & 0.092 & 0.236 & 0.284 & 0.092 & 0.700 & 0.951 & 0.047 & 8.018    & 9.178 & 0.363 \\
& \textbf{CHROM~\cite{de2013robust}}    & 0.222 & 0.281 & 0.070 & 0.222 & 0.281 & 0.070 & 0.671 & 1.054 & 0.105 & 7.229    & 8.922 & 0.512 \\
& \textbf{POS~\cite{wang2016algorithmic}}      & 0.236 & 0.286 & 0.136 & 0.236 & 0.286 & 0.136 & 0.652 & 0.954 & 0.135 & 7.354    & 8.040  & 0.492 \\
& \textbf{Rhythmnet~\cite{niu2019rhythmnet}}    & 0.163 & 0.185 & 0.166 & 0.163 & 0.185 & 0.166 & 0.242 & 0.329 & 0.278 & 6.023    & 7.131  & 0.707 \\
          & \textbf{Dual-GAN ~\cite{lu2021dual}}      & 0.189 & 0.092 & 0.157 & 0.189 & 0.092 & 0.157 & 0.411 & 0.366 & 0.243 & 6.552    & 7.621  & 0.668 \\
          & \textbf{PhysFormer++~\cite{yu2023physformer++}} & 0.084 & 0.113 & 0.182 & 0.084 & 0.113 & 0.182 & 0.323 & 0.365 & 0.306 & 5.635    & 7.258  & 0.769 \\
          & \textbf{HSRD-R~\cite{wang2023hierarchical}}       & 0.062 & 0.078 & 0.179 & 0.062 & 0.078 & 0.179 & 0.238 & 0.312 & 0.310 & 5.113    & 6.211  & 0.776 \\
          & \textbf{HSRD-T~\cite{wang2023hierarchical}}       & 0.173 & 0.267 & 0.166 & 0.173 & 0.267 & 0.166 & 0.321 & 0.558 & 0.264 & 6.176    & 7.344  & 0.672 \\
          & \textbf{Base-R~\cite{he2016deep}}    & 0.106 & 0.280 & 0.211 & 0.106 & 0.280 & 0.211 & 0.297 & 0.843 & 0.281 & 5.790    & 7.906  & 0.733 \\
          & \textbf{Base-T~\cite{wu2020visual}}          & 0.179 & 0.283 & 0.132 & 0.179 & 0.283 & 0.132 & 0.403 & 0.989 & 0.262 & 6.553    & 8.007  & 0.652 \\
& \textbf{PhysMLE-R}  & \textbf{0.056} & \textbf{0.073} & \textbf{0.248} & \textbf{0.056} & \textbf{0.073} & \textbf{0.248} & \textbf{0.223} & \textbf{0.296} & \textbf{0.345} & \textbf{4.353 }   & \textbf{5.081}  & \textbf{0.850} \\
& \textbf{PhysMLE-T}  & 0.063 & 0.082 & 0.201 & 0.063 & 0.082 & 0.201 & 0.233 & 0.301 & 0.315 & 4.927    & 5.746  & 0.831 \\
  \midrule                     
\textbf{BUAA} & \textbf{GREEN~\cite{verkruysse2008remote}}    & 0.347 & 0.395 & 0.087 & 0.347 & 0.395 & 0.087 & 0.645 & 0.863 & 0.092 & 5.823    & 7.988 & 0.562 \\
& \textbf{CHROM~\cite{de2013robust}}    & 0.379 & 0.324 & 0.068 & 0.379 & 0.324 & 0.068 & 0.681 & 0.884 & 0.072 & 6.093    & 8.294 & 0.517 \\
& \textbf{POS~\cite{wang2016algorithmic}}      & 0.320 & 0.376 & 0.096 & 0.320 & 0.376 & 0.096 & 0.628 & 0.842 & 0.113 & 5.041    & 7.120  & 0.637 \\
& \textbf{Rhythmnet~\cite{niu2019rhythmnet}}    & 0.242 & 0.389 & 0.277 & 0.242 & 0.273 & 0.277 & 0.634 & 0.891 & 0.274 & 4.630    & 8.102  & 0.774 \\
          & \textbf{Dual-GAN ~\cite{lu2021dual}}     & 0.218 & 0.385 & 0.281 & 0.218 & 0.385 & 0.281 & 0.626 & 0.860 & 0.263 & 4.309    & 7.871  & 0.784 \\
          & \textbf{PhysFormer++~\cite{yu2023physformer++}} & 0.201 & 0.394 & 0.289 & 0.201 & 0.394 & 0.289 & 0.599 & 0.926 & 0.282 & 4.021    & 8.711  & 0.790 \\
          & \textbf{HSRD-R~\cite{wang2023hierarchical}}       & 0.207 & 0.371 & 0.289 & 0.207 & 0.371 & 0.289 & 0.612 & 0.879 & 0.286 & 4.176    & 7.502  & 0.814 \\
          & \textbf{HSRD-T~\cite{wang2023hierarchical}}       & 0.189 & 0.172 & 0.313 & 0.189 & 0.172 & 0.313 & 0.583 & 0.697 & 0.320 & 3.073    & 6.912  & 0.866 \\
          & \textbf{Base-R~\cite{he2016deep}}    & 0.324 & 0.411 & 0.269 & 0.324 & 0.411 & 0.269 & 0.642 & 0.917 & 0.269 & 4.914    & 8.614  & 0.758 \\
          & \textbf{Base-T~\cite{wu2020visual}}          & 0.385 & 0.498 & 0.195 & 0.385 & 0.498 & 0.195 & 0.723 & 0.989 & 0.143 & 5.672    & 10.490 & 0.679 \\
& \textbf{PhysMLE-R}  & \textbf{0.131} & \textbf{0.156} & \textbf{0.362} & \textbf{0.131} & \textbf{0.156} & \textbf{0.362} & \textbf{0.501} & \textbf{0.635} & \textbf{0.437} & \textbf{2.162}    & \textbf{3.580 } & \textbf{0.922} \\
& \textbf{PhysMLE-T}  & 0.148 & 0.169 & 0.291 & 0.148 & 0.169 & 0.291 & 0.562 & 0.698 & 0.299 & 3.456    & 7.038  & 0.834 \\
\bottomrule 
\end{tabular}
\vspace{-2mm}
\vspace{-2mm}
\end{table*}

\subsection{Comparison Experiment over Protocol I}
In this section, we present and discuss the evaluation results based on six datasets following Protocol I. Notes that, for single-task baselines, only one type of target label was used for training. The Base-R/T is a multi-task variant that is built by one backbone network with four task estimation heads like PhysMLE, and all labels were used for training regardless of the types of labels of the domains. Those compared DG methods were also constructed based on the Base structures.  

\subsubsection{Multi-task evaluation}
Firstly, we present results when PURE and VIPL-HR are target domains (as shown in Table \ref{t2}), and compared them against multiple baseline methods, including single-task methods and DG methods. It indicates that although traditional SpO2 methods \cite{tarassenko2014non} can compete with some single-task DL-based method \cite{akamatsu2023blood} and DG methods \cite{ganin2015unsupervised, krueger2021out, parascandolo2020learning} integrated with ResNet-18 on relatively simple scenarios (i.e., PURE), DG methods mostly outperformed all traditional approaches in both HR and SpO2 estimation. This suggests traditional single-task approaches struggle to handle intricate motions and scenarios compared to deep learning. Referring to Table \ref{t3} that contains results of V4V and HCW, state-of-the-art (SOTA) DL-based multi-task methods \cite{liu2020multi, narayanswamy2024bigsmall} did not outperform single-task traditional method \cite{tarassenko2014non} in RR estimation notably.

Nevertheless, as a multi-task model, PhysMLE still outperformed all baselines across all tasks. It shows that PhysMLE can achieve a promising balance among different tasks. Besides, some interesting findings were identified when we integrated different backbones into DG methods and our proposals. We notice that, different backbone networks led to performance disparity even when combined with the same method. Particularly, despite PhysMLE-R performing better than PhysMLE-T in most cases, PhysMLE-T led to higher accuracy in the SpO2 estimation compared to PhysMLE-R. A similar phenomenon was observed for those DG approaches. Thus, we assume that the ViT architecture is better at handling complex transformations between facial color signals and SpO2 values than CNN. Additionally, we found not all DG methods can be comparable to the SOTA single-task method \cite{yu2023physformer++} in HR estimation. It suggests that previous DG methods, even for those designed for rPPG \cite{lu2023neuron, wang2023hierarchical}, are still not good at imbalance learning of multiple tasks. Lastly, soft-parameter sharing PhysMLE performed notably better than the multi-task base models that followed the hard-parameter sharing paradigm and performed comparatively to multi-task baselines \cite{liu2020multi, narayanswamy2024bigsmall}. The computational cost analysis is provided in the following sections.

\subsubsection{HRV evaluation}
For two single-task datasets (i.e., UBFC-rPPG and BUAA), we used the HRV index (LFnu, HFnu, and LF/HF) and HR evaluation to compare the quality of the predicted BVP signal. As shown in Table \ref{t4}, similar findings from the Multi-task evaluation section are also identified in this part. Particularly, compared to single-task baselines, getting trained in multi-task labels did not bring negative effects to PhysMLE, which still notably outperformed traditional methods and the previous single-task baseline \cite{niu2019rhythmnet, lu2021dual, yu2023physformer++}. This suggests that PhysMLE can explore shared information across multiple task labels effectively to enhance generalizability in a single-task target domain. Additionally, the best results of PhysMLE in all HRV metrics underscore the capability of PhysMLE to yield high-quality LF and HF (Hz) prediction. This can be beneficial to applications in emotion recognition, and healthcare \cite{gouveia2023low,wu2023recognizing}.

\begin{figure}
\begin{center}
\includegraphics[scale=0.25]{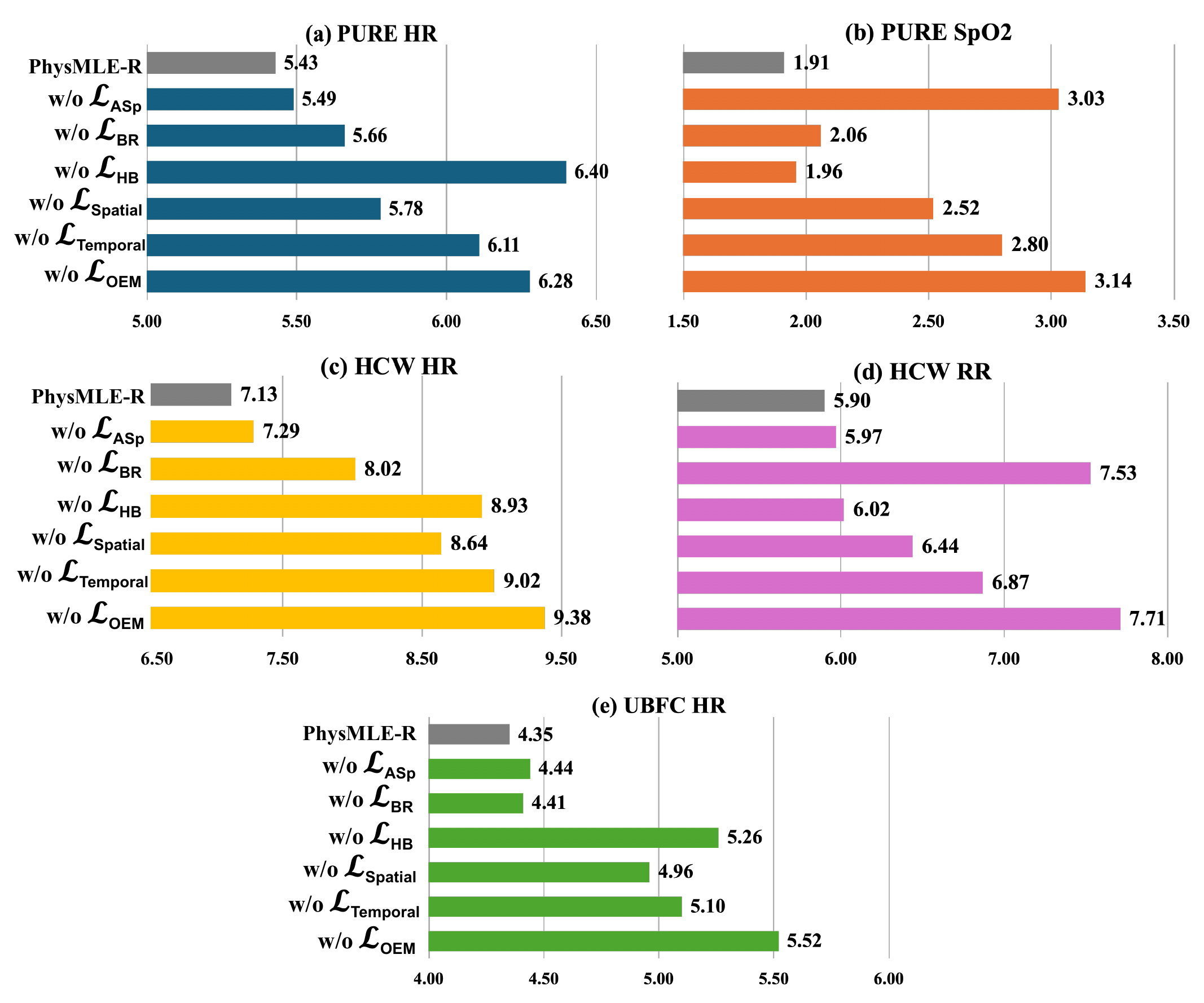}\\
\end{center}
\vspace{-6mm}
\caption{MAE of applying different regularizations over PhysMLE-R.}\label{f7}
\vspace{-6mm}
\end{figure}

\subsection{Ablation Study over Protocol I}
\subsubsection{Impact of regularizations}
In Figure \ref{f7}, the MAE of PhysMLE-R and its variants on three datasets (i.e., PURE, HCW, and UBFC-rPPG) over Protocol I are presented to illustrate the effect of each tailored regularization in the MSSDG challenge. Overall, missing any of the regularizations results in performance degradation compared to the full PhysMLE. Specifically, we first notice that the variant without $\mathcal{L}_{OEM}$ presents a notable performance decline compared to PhysMLE-R. We argue that the lack of constraints on the distinct parameters learned by experts led to the tendency of experts under the MoE architecture to learn approximate and easy-to-learn features, which degraded the model into a multi-head-like structure. Referring to Table \ref{t2} to \ref{t4}, it was found that the performance of PhysMLE without $\mathcal{L}_{OEM}$ is only slightly better than that of Base-R. Besides, we found that the $\mathcal{L}_{HB}$ mainly contributed to the HR prediction tasks, despite it still helping the learning of other tasks slightly. Similarly, the positive effects of $\mathcal{L}_{BR}$ and $\mathcal{L}_{ASp}$ were identified in RR and SpO2 estimation, respectively. It illustrates the effectiveness of our prior-based regularization design, and the enhanced task-specific feature can help alleviate the task imbalance problem under MSSDG.

\begin{table}[!t] 
\setlength{\tabcolsep}{1mm}
\scriptsize
\centering
\caption{Results of Variants with Different Hyper-parameter $\alpha$ and $r$ on MSSDG Protocol I.}
\label{t5}
\begin{tabular}{cccccccccc} 
\toprule  
& & \multicolumn{4}{c}{\textbf{PURE}}             & \multicolumn{4}{c}{\textbf{HCW}} \\
\cmidrule(lr){3-6} \cmidrule(lr){7-10} 
\multicolumn{2}{c}{\textbf{Hyper}} & \multicolumn{2}{c}{\textbf{HR}}& \multicolumn{2}{c}{\textbf{SpO2}} & \multicolumn{2}{c}{\textbf{HR}}& \multicolumn{2}{c}{\textbf{RR}} \\
 \cmidrule(lr){3-4} \cmidrule(lr){5-6} \cmidrule(lr){7-8} \cmidrule(lr){9-10}
 \textbf{$\mathbf{\alpha}$}& \textbf{$\mathbf{r}$} &  \textbf{MAE↓}& \textbf{\quad p↑ \quad} & \textbf{MAE↓}& \textbf{\quad p↑ \quad}& \textbf{MAE↓}& \textbf{\quad p↑ \quad}& \textbf{MAE↓}& \textbf{\quad p↑ \quad} \\
  \midrule 
  \textbf{8}  & \textbf{8}  & 10.40         & 0.46          & 2.89          & 0.20          & 12.97         & 0.23          & 6.13          & 0.06           \\
   & \textbf{16} & 9.00          & 0.52          & 4.17          & 0.17          & 9.66          & 0.44          & 7.41          & 0.03           \\
   & \textbf{32} & 8.03          & 0.56          & 6.14          & 0.11          & 7.79          & 0.59          & 8.21          & 0.01           \\
\textbf{16} & \textbf{8}  & 7.13          & 0.63          & 3.49          & 0.19          & 9.61          & 0.41          & 6.10          & 0.07           \\
   & \textbf{16} & 6.69          & 0.67          & 5.91          & 0.13          & 10.12         & 0.34          & 5.99          & 0.09           \\
   & \textbf{32} & 6.80          & 0.66          & 5.77          & 0.13          & 10.89         & 0.31          & 5.96          & 0.09           \\
\textbf{32} & \textbf{8}  & 9.48          & 0.50          & 3.21          & 0.19          & 8.10          & 0.48          & 5.93          & 0.10           \\
   & \textbf{16} & \textbf{5.43} & \textbf{0.77} & \textbf{2.19} & \textbf{0.22} & \textbf{7.13} & \textbf{0.61} & \textbf{5.90} & \textbf{0.10}  \\
   & \textbf{32} & 6.67          & 0.67          & 3.74          & 0.18          & 7.49          & 0.54          & 6.15          & 0.06 \\
\bottomrule 
\end{tabular} 
\vspace{-1.4mm}
\end{table}

\begin{figure}
\begin{center}
\includegraphics[scale=0.38]{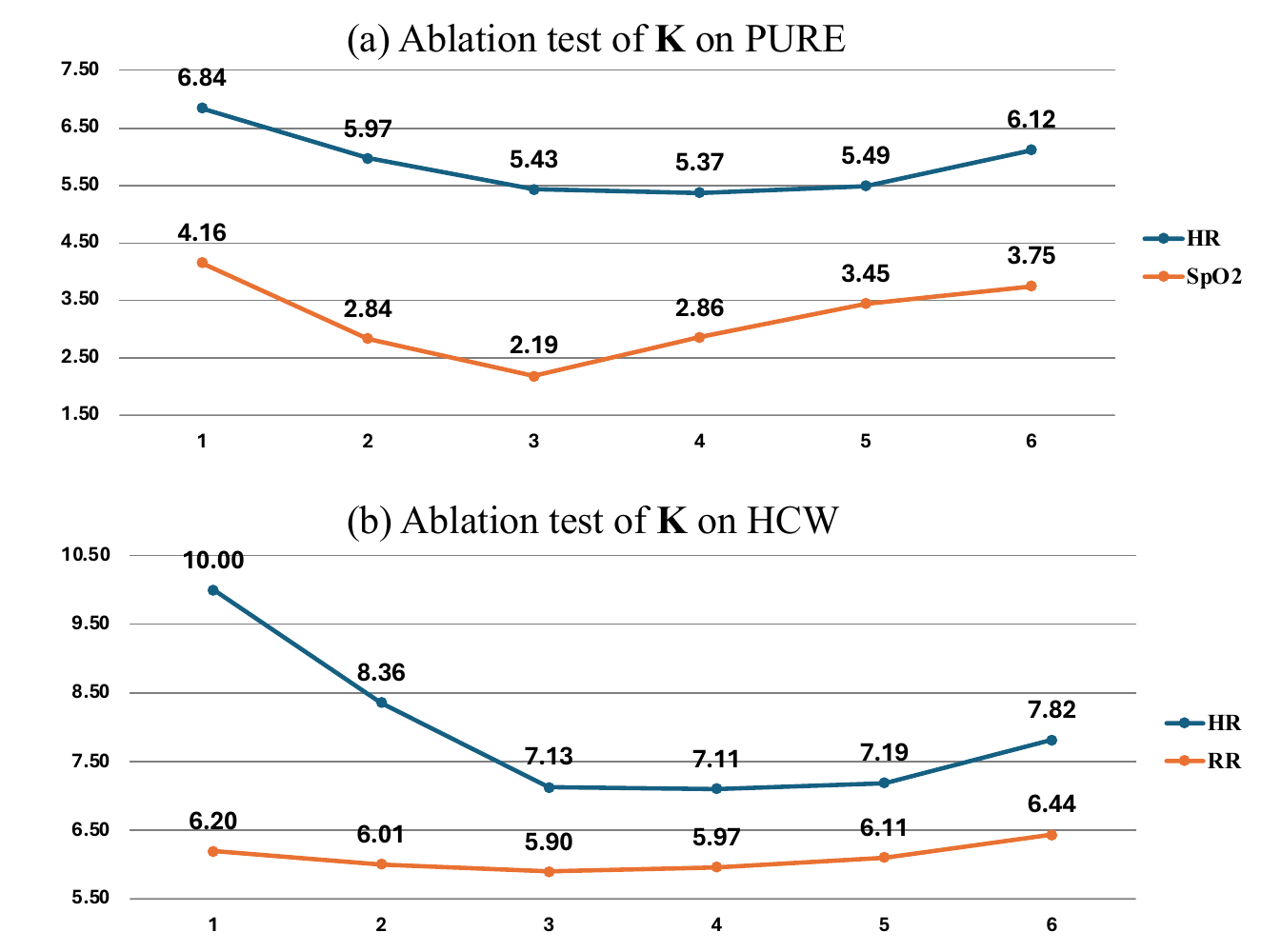}\\
\end{center}
\vspace{-6mm}
\caption{MAE of PhysMLE-R with different numbers of experts.}\label{f8}
\end{figure}

\begin{table}[!t] 
\setlength{\tabcolsep}{1mm}
\scriptsize
\centering
\caption{Computational cost of variants with different structures on MSSDG Protocol I.}
\label{t6}
\begin{tabular}{ccccccc} 
\toprule  
& \multicolumn{2}{c}{\textbf{PURE}}             & \multicolumn{2}{c}{\textbf{HCW}} & & \\
\cmidrule(lr){2-3} \cmidrule(lr){4-5} 
& \textbf{HR}   & \textbf{SpO2}              & \textbf{HR}    & \textbf{RR}              & \textbf{Trainable Params(M)} & \textbf{FLOPs(G)}  \\
\midrule
Base-R$^{*}$       & 8.74 & 3.93              & 10.18 & 8.22            & 13.81                & 7.32      \\
MoE-R$^{*}$      & 5.45 & 2.77              & \textbf{7.02}  & 6.08            & 44.79                & 58.92     \\
Softmax-R$^{+}$ & 6.95 & 3.14              & 8.16  & 7.60            & 10.61                 & 22.16      \\
PhysMLE-R$^{+}$    & \textbf{5.43} & 2.19              & 7.13  & \textbf{5.90}            & 13.85                 & 34.57      \\
\midrule
Base-T$^{*}$       & 8.81 & 2.45              & 10.24 & 9.74            & 82.57                & 15.70     \\
MoE-T$^{*}$        & 6.84 & 2.01              & 7.50  & 6.62            & 214.42               & 90.07    \\
Softmax-T$^{+}$& 7.20 & 3.36              & 8.13  & 8.33            & 69.66                & 46.89     \\
PhysMLE-T$^{+}$    & 6.62 & \textbf{1.86}              & 7.34  & 6.55            & 82.90                & 69.00 \\
\bottomrule 
\end{tabular} 
\textnormal{\\Notes: In this table, the MAE of each task is provided. The $^{+}$ means that its experts are based on LoRA, and the $^{*}$ means full fine-tuning.}
\vspace{-1.4mm}
\end{table}

\subsubsection{Impact of hyperparameters}
As shown in Table \ref{t5}, given various combinations of $\alpha$ and $r$, we present the MAE and $p$ of the multi-task evaluation over PURE and HCW. Generally, we notice that the $r$ was positively associated with the capacity of HR estimation and negatively related to other tasks (i.e., SpO2 and RR). We assume it is because $r$ determines the rank of each expert's parameters, which is crucial to the number of learned features. Compared to HR, the range of SpO2 and RR is narrower, which means that there are fewer and simpler mapping relationships between the inputted STMaps and the predicted values. Thus, for multi-task physiological measurement, too high or too low $r$ can lead to a severe see-saw effect between tasks in the model. Additionally, for the effect of $\alpha$ solely, although there is no clear pattern to its effects on the model performance, some interesting findings are elaborated for the effect of $\frac{\alpha}{r}$. As stated in \cite{hu2021lora}, $\alpha$ is used to scale the weights of LoRAs. Therefore, we find that when $r$ was 8 or 32, too large $\frac{\alpha}{r}$ can amplify the seesaw effect; when $\frac{\alpha}{r}$ was too small, even though $r$ was chosen to be a more appropriate value of 16, the performance of the model can be restricted by the fact that the features learned by LoRA experts were overly narrowed. Overall, the choice of appropriate $r$ and $\alpha$ has a crucial impact on the performance of a LoRA-based model.

\begin{table*}[t]
\setlength{\tabcolsep}{1mm}
\centering
\scriptsize
\caption{Comparison Experiment Results on MSSDG Protocol II.}
\begin{tabular}{cccccccccccccc}
\toprule 
                          & & \multicolumn{6}{c}{\textbf{VIPL-HR}}             & \multicolumn{6}{c}{\textbf{HCW}}\\
\cmidrule(lr){3-8} \cmidrule(lr){9-14} 
 &      & \multicolumn{3}{c}{\textbf{HR}}    & \multicolumn{3}{c}{\textbf{SpO2}}   &     \multicolumn{3}{c}{\textbf{HR}}    & \multicolumn{3}{c}{\textbf{RR}}     \\
                    \textbf{Method Type}             & \textbf{Method} & \textbf{MAE↓}  & \textbf{RMSE↓} & \textbf{\quad p↑ \quad}    & \textbf{MAE↓}  & \textbf{RMSE↓} & \textbf{\quad p↑ \quad}    & \textbf{MAE↓}  & \textbf{RMSE↓} & \textbf{\quad p↑ \quad}    & \textbf{MAE↓}  & \textbf{RMSE↓}  & \textbf{\quad p↑ \quad}    \\
                    \midrule 
\textbf{Single-task} & \textbf{GREEN \cite{verkruysse2008remote} }       & 12.18 & 18.23 & 0.25   & —    & —     & —  & 14.36 & 20.01 & 0.04 & —    & —    & —   \\
\textbf{Traditional}& \textbf{CHROM \cite{de2013robust}  }     & 11.44 & 16.97 & 0.28   & —    & —     & —  & 14.66 & 19.81 & 0.05 & —    & —    & —      \\
 & \textbf{POS \cite{wang2016algorithmic} }         & 14.59 & 21.26 & 0.19   & —    & —     & — & 12.37 & 17.61 & 0.11 & —    & —    & —     \\
                        & \textbf{ARM-SpO2 \cite{tarassenko2014non}}           & —     & —     & —      & 8.11 & 12.23 & 0.08     & —    & —     & —       & —    & —     & —  \\
                        & \textbf{ARM-RR \cite{tarassenko2014non}}           & —     & —     & —      & —    & —     & —        & —    & —     & —       & 8.08 & 8.92  & 0.01 \\
                        \midrule 
\textbf{Single-task}        & \textbf{RhythmNet \cite{niu2019rhythmnet} }   & 5.30  & 8.14  & 0.76   & —    & —     & —        & 6.27 & 10.06 & 0.65    & —    & —     & — \\
\textbf{Deep} & \textbf{BVPNet \cite{das2021bvpnet} }     & 5.34  & 7.85  & 0.70   & —    & —     & —        & 6.51 & 10.22 & 0.64    & —    & —     & —   \\
                        & \textbf{Dual-GAN \cite{lu2021dual} }    & 4.93  & 7.68  & 0.81   & —    & —     & —        & 5.33 & 9.87  & 0.69    & —    & —     & —   \\
                        & \textbf{PhysFormer++ \cite{yu2023physformer++}} & 4.88  & 7.62  & 0.80   & —    & —     & —        & 4.88 & 9.01  & 0.72    & —    & —     & —  \\
                        & \textbf{EfficientPhys \cite{liu2023efficientphys}} & 5.84  & 8.91  & 0.72   & —    & —     & —        & 6.90 & 11.13 & 0.62    & —    & —     & —   \\
                        & \textbf{rPPG-MAE$^{*}$ \cite{liu2024rppg} }    & 4.52  & 7.56  & 0.81   & —    & —     & —        & 4.11 & 8.20  & 0.78    & —    & —     & —          \\
                        & \textbf{Contrast-Phys+$^{*}$ \cite{sun2024}} & \textbf{4.06}  & \textbf{7.10}  & \textbf{0.83}   & —    & —     & —        & \textbf{3.83} & \textbf{7.57}  & \textbf{0.82}    & —    & —     & —   \\
                        & \textbf{rSPO \cite{akamatsu2023blood}}         & —     & —     & —      & 2.36 & 6.54  & 0.16     & —    & —     & —       & —    & —     & — \\
                        \midrule 
\textbf{Multi-task} & \textbf{MTTS-CAN \cite{liu2020multi} }     & —     & —     & —      & —    & —     & —        & 8.22 & 13.31 & 0.51    & 7.66 & 11.17 & 0.04   \\
\textbf{Deep} & \textbf{BigSmall \cite{narayanswamy2024bigsmall}}    & —     & —     & —      & —    & —     & —        & 7.10 & 11.64 & 0.58    & 5.42 & 9.53 & 0.06   \\
\midrule 
\textbf{Ours} & \textbf{Base-R \cite{he2016deep} }    & 4.64  & 7.98  & 0.80   & 2.59 & 6.77  & 0.12     & 4.94 & 9.07  & 0.71    & 4.65 & 7.81  & 0.08 \\
                        & \textbf{Base-T \cite{wu2020visual} }        & 4.91  & 8.57  & 0.76   & 1.64 & 5.78  & 0.19     & 5.39 & 9.87  & 0.68    & 5.03 & 8.60  & 0.07 \\
                       & \textbf{PhysMLE-R }    & 4.29  & 7.37  & 0.81   & 1.85 & 5.92  & 0.17     & 3.88 & 7.63  & 0.81    & \textbf{2.23} & \textbf{4.11}  & \textbf{0.14} \\
                        & \textbf{PhysMLE-T}     & 4.79  & 8.06  & 0.79   & \textbf{1.55} & \textbf{5.40}  & \textbf{0.21}     & 4.77 & 8.90  & 0.75    & 3.14 & 5.28  & 0.12\\
\bottomrule 
\end{tabular}
\textnormal{\\ Notes: In this table, $^{*}$ means it has been self-supervised pre-trained over the target dataset before training.}
\label{t7}
\end{table*}

\subsubsection{Impact of the number of experts}
We further conducted ablation experiments for PhysMLE-R with different numbers $K$ of experts over PURE and HCW. The $K$ was adjusted from 1 to 6, and the MAE was presented in Figure \ref{f8}.

As shown by the blue curves in Figure \ref{f8}, we notice that the best HR estimation performance was acquired by the variant with 4 experts. The performance curve of HR was also overall steady when $K$ was between 3 and 5. This is because the four physiological estimation tasks can be implicitly aligned with the experts by the expert orthogonality $\mathcal{L}_{OEM}$. Since HR estimation is highly correlated with the BVP estimation task and most of the datasets in MSSDG provide HR labels computed from the BVP signals, a model with 4 experts had more experts to learn HR-related features compared to a model with 3 experts. We call this task as strong task, which means that they share more features internally and have a relatively balanced number of labels available for training in cross-domain scenarios.

However, when we look at the orange curves for physiological signals such as SpO2 and RR, which were relatively weak tasks (i.e., less internal correlation and training labels), we noticed that more experts do not bring significant benefit to such tasks except for the fact that they bring more computational complexity (FLOPs(G) of 53.43 for K=4, much higher than 34.57 for K=3). Especially, for SpO2, it is essentially not directly accessible from the BVP signal and the curve of SpO2 showed a significant increase when $K$ exceeded 3. This means that the model appears to have a more obvious seesaw effect. In summary, we chose $K=3$ experts as the best parameter for PhysMLE in our experiment.

\subsubsection{Impact of key component}
In this section, we validate the effectiveness and computational costs of each key component of the PhysMLE. For comparison, we set up three types of baselines: (1) the Base model that solely consists of one backbone network and multi-task heads with full fine-tuning; (2) the MoE model that replaces the $K=3$ LoRA experts in PhysMLE with the two full parameter experts and the original weights of the backbone network; (3) the Softmax model that remains the LoRA experts but replaces EFRouter in PhysMLE with softmax gating \cite{shazeer2017outrageously}. Except for the MAE of each estimation task, we provide the number of trainable parameters and FLOPs as well. 

\begin{figure*}
\begin{center}
\includegraphics[scale=0.5]{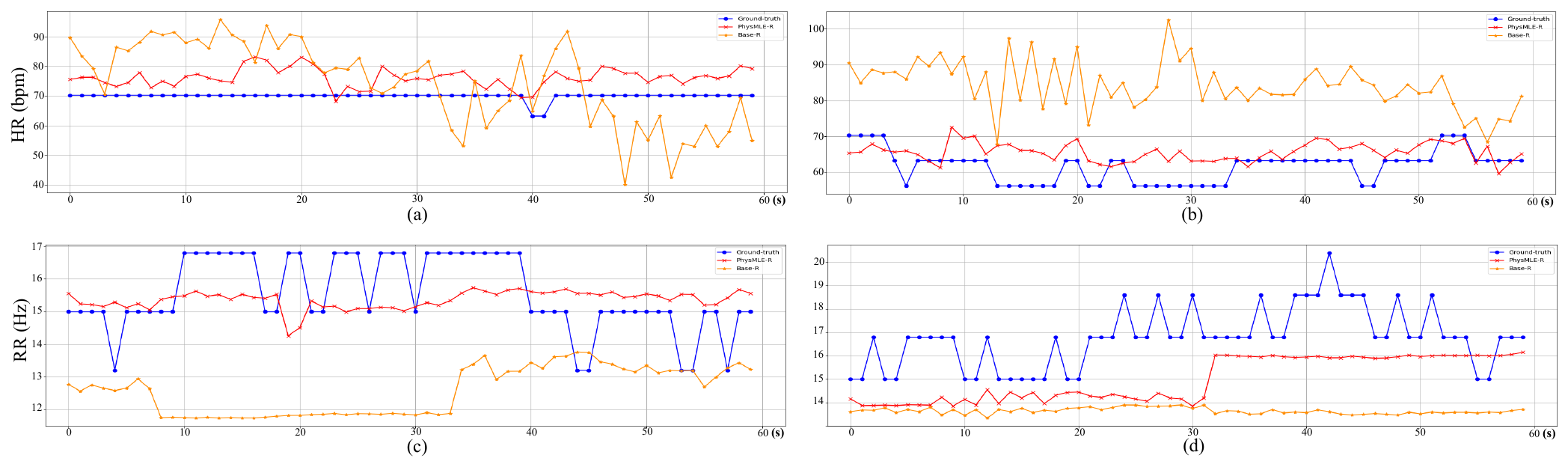}\\
\end{center}
\vspace{-6mm}
\caption{Visualization of predicted physiological signals on HCW. The \textcolor{blue}{BLUE} line indicates the ground-truth signal, the \textcolor{red}{RED} line indicates the predicted signal by PhysMLE, and the \textcolor{orange}{ORANGE} line is the predicted signal by Base.}\label{f9}
\end{figure*}

From the results of Table \ref{t6}, we find that the PhysMLE generally performed well on most tasks. Besides, in terms of computational cost, only the Softmax models needed a smaller number of trainable parameters and thus lower arithmetic consumption. However, the performance of Softmax was only better than the Base models and much worse than PhysMLE. Thus, although our proposed EFRouter introduces additional parameters and computational effort, the significant performance improvement shows that it is still a relatively promising option for the gating mechanism. Compared to the MoE methods, the full fine-tuning brought a notable increase in the number of training parameters and FLOPs. However, on the more complex dataset of HCW, the performance of full fine-tuning methods in HR measurement was slightly better than PhysMLE, and in other cases, the performance was close to our proposal as well; it indicates that the HR estimation task in complex scenarios still requires a certain number of parameters and model complexity to realize. Nevertheless, considering the overall better performance of PhysMLE and the relatively acceptable training cost (training parameters number is very close to Base), we believe that PhysMLE is competitive.

\subsection{Comparison Experiment over Protocol II}

\subsubsection{Intra-dataset test}
In this section, considering the diverse illumination, motion scenes, and acquisition devices in VIPL-HR and HCW, we provide the results of the five-fold cross-validation experiment over these two challengable datasets in Table \ref{t7}. We found that only the SOTA self-supervised pre-train single-task model \cite{liu2024rppg} outperformed our proposal in HR estimation. Nevertheless, PhysMLE showed a comparatively better performance than other single-task methods, including one pre-trained method \cite{liu2024rppg} in the single-task indicator. At the same time, PhysMLE enabled simultaneous multiple human vital signs measurements with comparable or better performance than baseline models. Compared to multi-task methods \cite{liu2020multi, narayanswamy2024bigsmall}, our proposal still kept the leading performance. In all, the results further supported the effectiveness of our approach in addressing the challenges
posed by these two complex datasets.

\begin{figure*}
\begin{center}
\includegraphics[scale=0.5]{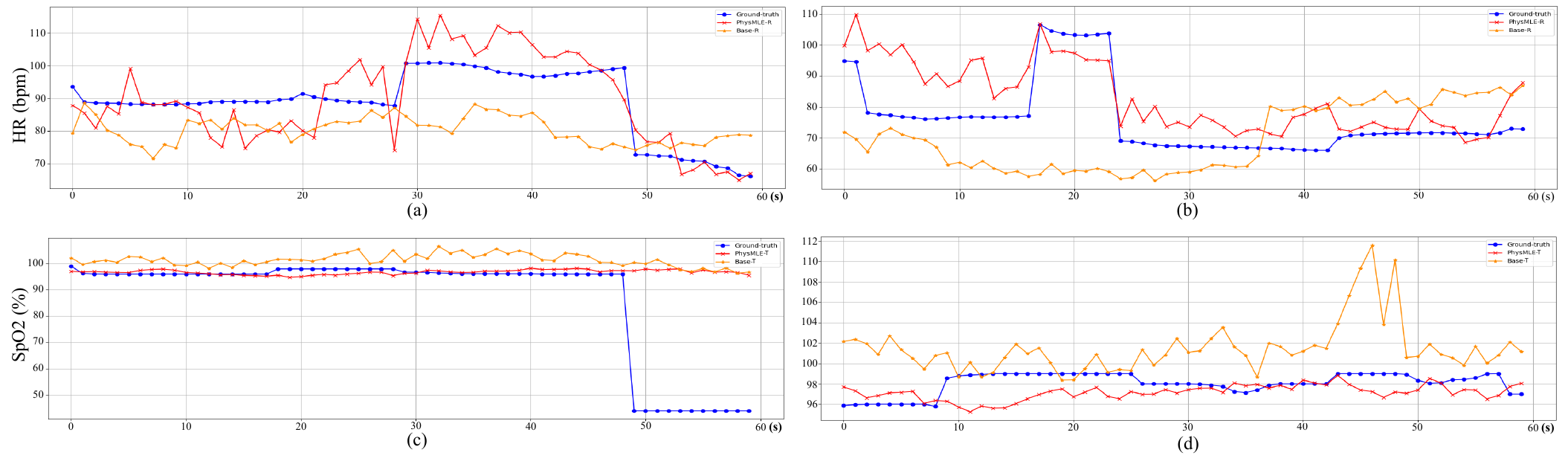}\\
\end{center}
\vspace{-6mm}
\caption{Visualization of predicted physiological signals on VIPL-HR. The \textcolor{blue}{BLUE} line indicates the ground-truth signal, the \textcolor{red}{RED} line indicates the predicted signal by PhysMLE, and the \textcolor{orange}{ORANGE} line is the predicted signal by Base.}\label{f10}
\end{figure*}

\begin{figure}
\begin{center}
\includegraphics[scale=0.19]{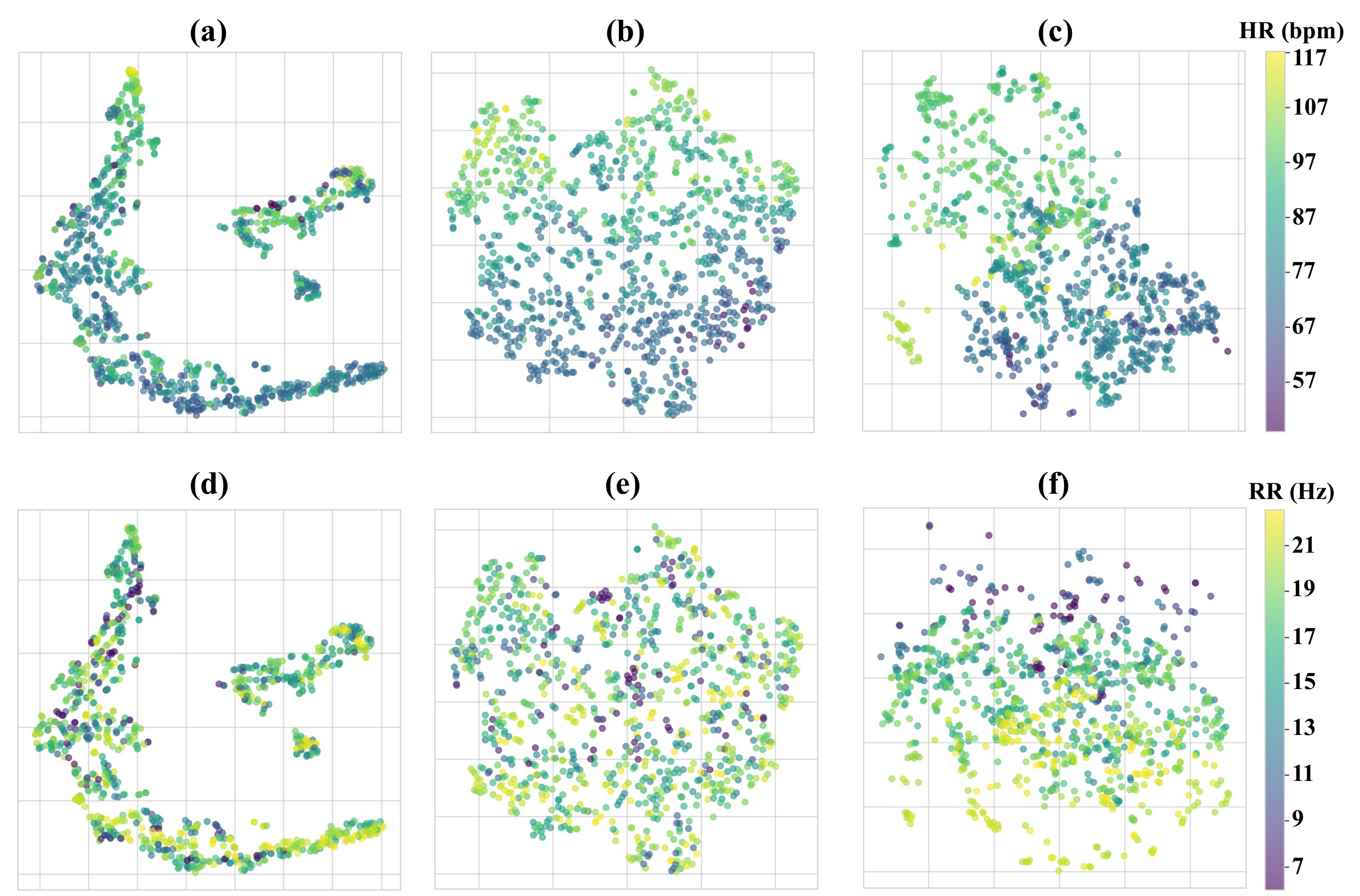}\\
\end{center}
\vspace{-6mm}
\caption{Visualization of the extracted features on HCW under Protocol II. The subfigure (a)(d) is the feature produced by Base-R, and the others are given by PhysMLE-R. Among them,(b)(e) reflects the task-agnostic feature $s^{'}$, and (c)(f) is the task-specific feature space after the EFRouter. The HR and RR value is represented as the color lightens.}\label{f11}
\end{figure}

\begin{figure}
\begin{center}
\includegraphics[scale=0.19]{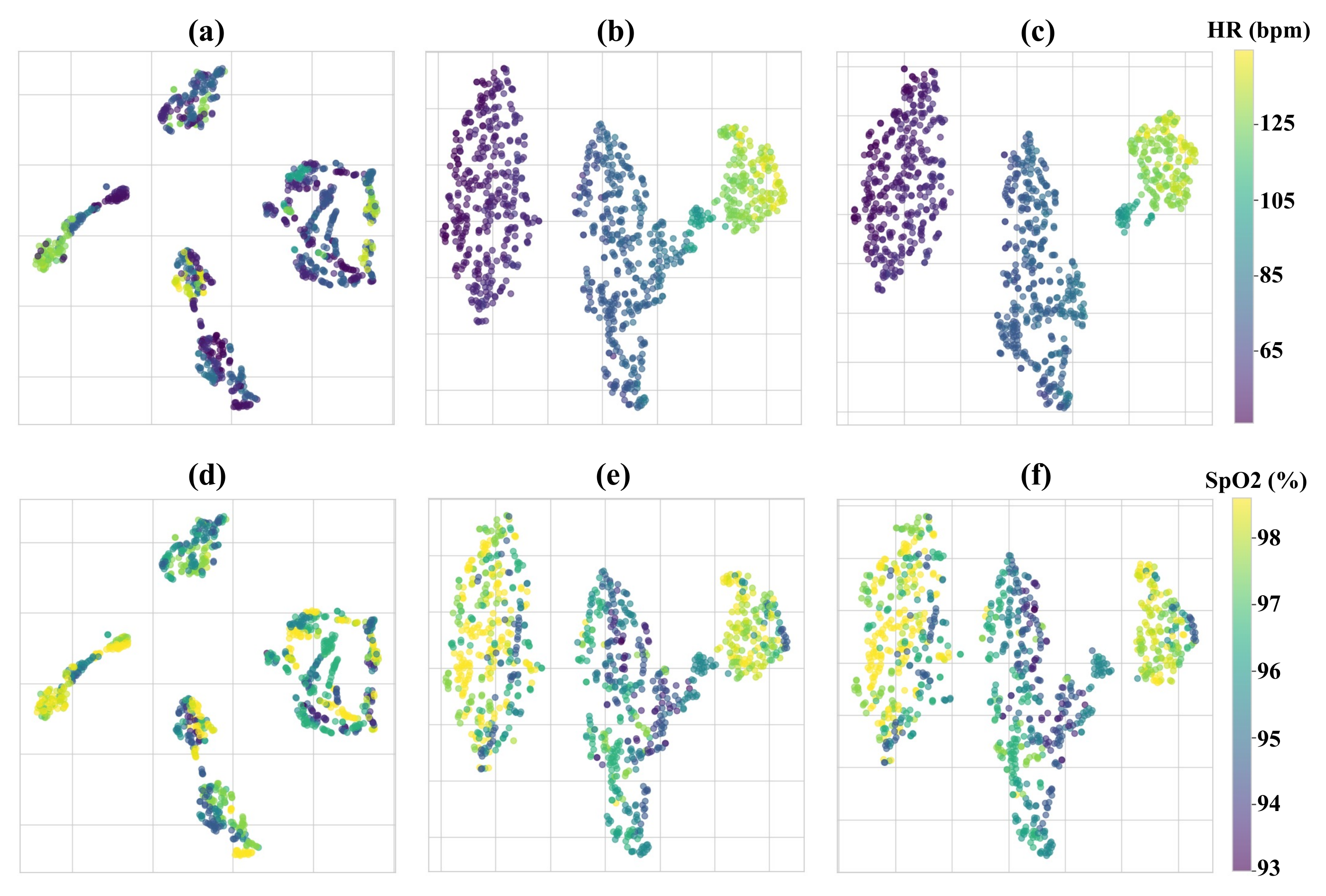}\\
\end{center}
\vspace{-6mm}
\caption{Visualization of the extracted features on VIPL-HR under Protocol II. The subfigure (a)(d) is the feature produced by Base-R, and the others are given by PhysMLE-R. Among them, (b)(e) reflects the task-agnostic feature $s^{'}$, and (c)(f) is the task-specific feature space after the EFRouter. The HR and SpO2 value is represented as the color lightens.}\label{f12}
\end{figure}

\subsubsection{Visualization of predicted vital signs}
We present and visualize some example predictions (about 1 minute) from both HCW and VIPL-HR. Firstly, as shown in Figure \ref{f9}(a)(b), we found that the output HR of the PhysMLE-R was overall smoother and closer to ground-truth compared to Base-R. We believe this was benefited from our temporal-consistent constraints $\mathcal{L}_{Temporal}$. By contrast, for VIPL-HR, there are more HR mutations in the provided ground-truth HR (e.g., HR at position 28s in Figure \ref{f10}(a) mutates from around 90 to 100). As a result, the HR curve inference by PhysMLE trained on VIPL-HR also showed more fluctuation when compared to that on HCW, while numerically and in terms of the overall trend, it was still closer to ground truth compared to Base-R.

As for other tasks, for RR estimation in HCW, we found the output RR curve of our proposal is smoother and basically closer to the mean value of ground truth compared to Base-R. However, we also found that individually occurring RR fluctuations (e.g., in Figure \ref{f9}(c), RR should rise to 17 Hz overall between 10s and 40s) were not captured by our PhysMLE-R. From Figure \ref{f9}(d), we also note that the overall RR rise that should occur around the 20th second was not perceived by PhysMLE-R until the 30th second. These indicate that our model, as a multi-task model, still has room for improvement in capturing RR transformations more accurately. We expect future work to further address this issue. Nevertheless, on the SpO2 estimation task, our model showed a competitive performance. Specifically, refering to Figure \ref{f10}(c)(d), although the curve of Base-T and PhysMLE-T was close, our proposal presented higher stability, and its output was more in line with the characteristics of SpO2. For instance, we noticed that the output SpO2 value of Base-T sometimes exceeded 100 (\%), which should be the hard range boundary of SpO2. By contrast, the output of PhysMLE-T always remained in the reasonable numeric range, which we assume benefited from our $\mathcal{L}_{Asp}$. At the same time, when there were abnormal values in the ground truth (e.g., SpO2 should not change suddenly from around 95 to below 50 in Figure \ref{f10}(c)), PhysMLE-T could still stabilize its output. This suggests that PhysMLE is capable of performing the SpO2 estimation task in a stable and robust manner.

\subsubsection{Visualization of feature space}
Based on protocol II, we randomly select 1000 samples from the test set and visualize their feature spaces by t-SNE \cite{van2008visualizing}. Firstly, see subfigures (a)(d) in Figure \ref{f11}, \ref{f12}, which represent the feature space of Base-R after training. As one type of hard-parameter sharing structure, the features encoded by the backbone network are shared among tasks and directly forwarded to each estimation head. Thus, we identify no clear distribution pattern or classification plane between samples with different physiological signs. Due to the diversity of physiological metrics and the specificity between physiological metrics, constructing the same feature space for different tasks for the same sample may exacerbate the negative effect on task-wise performance. This finding can also be identified from subfigures (b)(e) in Figure \ref{f11}, \ref{f12}. These spaces are the task-agnostic feature $s^{'}$ encoded by the PhysMLE-R before feeding to the task-specific heads and routers. We notice that, for the task with abundant training labels (i.e., HR), there is a clear classification plane (Figure \ref{f12}(b)) or hierarchical distribution with increasing HR (Figure \ref{f11}(b)). By contrast, for those weak tasks that are with limited labels, there are no very clear identifiable distributional differences between samples with different levels of RR or SpO2. It suggests that, a single and uniform feature space is not appropriate for multi-task physiological measurement, especially in the MSSDG setting, where weak tasks tend to be compromised by learning from strong tasks.

Refer to the subfigures (c)(f) in Figure \ref{f11}, \ref{f12}, we find that, with task-specific EFRouter feature selection for $s^{'}$, there is a clearer pattern of sample distribution as the level of physiological indicators changes, and there are disparities between the feature spaces of the same batch of samples under different tasks. Therefore, constructing new feature spaces for different tasks is important for multi-task learning.

\section{Limitation}
Through our extensive experiments, although it is demonstrated that our method remains quite competitive on both MSSDG and classic intra-dataset multi-task evaluations, there are still some limitations and pending issues to be addressed. The first one is on the number of parameters.  Due to our proposed EFRouter, as an indispensable component in the forward process of the model, it is not possible to merge the weights directly on the frozen weights after fine-tuning by utilizing LoRA as in previous works \cite{hu2021lora}. Therefore, although our model reduces the training cost of experts significantly by LoRA during the training process, our model always needs to load both the experts and the full backbone parameters. While we believe this is an unavoidable problem due to the MoE architecture, we still hope that future work can address it.

In addition, the accuracy of the RR detection remains somewhat flawed even in the intra-dataset setting (e.g., not being able to sense changes in RR in time). We believe that this is a common problem for rPPG-based RR detection. The basic principle of extracting respiration rate from rPPG is based on HRV, as we mentioned before in Priors for RR, and this method is, in fact, still interfered with by irregular respiration patterns, individual variability, and illumination, etc. In this paper, as a multi-task model, we only rely on the rPPG-based approach in order to extract more shared features among tasks. Future multi-task remote physiological monitoring work should consider more human movement features during respiration for RR estimation.

\section{Conclusion}
In this paper, we propose a low-rank MoE-based multi-task remote physiological measurement method PhysMLE. In terms of structural design, we innovatively fuse MoE and LoRA to address multi-task settings to learn shared and unique features in a more efficient way. Our proposed EFRouter fuses and selects features to serve different estimation tasks more accurately than the gating mechanism used in previous MoE approaches. At the same time, considering the challenge of domain shift between datasets and imbalance label space between tasks, which are often encountered in the real world, we define the MSSDG protocol and provide a larger scale benchmark. To fulfill our protocol, in addition to five public datasets that have been widely used in previous work, we collected a new dataset HCW to consider physiological monitoring in high cognitive workload states. In terms of algorithmic innovations, in addition to spatio-temporal semantic consistency being used to learn shared features, we innovatively introduce and instantiate three prior knowledge to solve the issue of unbalanced task labeling under MSSDG. Through experiments on various datasets, extensive baseline comparisons, structure combination, and hyperparameter selection, we demonstrate the effectiveness of our proposed method on multiple evaluation protocols and multiple physiological monitoring tasks. Future works are expected to make efforts on more efficient multi-task modeling structures and more accurate RR estimation.

\bibliographystyle{IEEEtran}
\bibliography{IEEEabrv,refs}

\end{document}